\documentclass{article}
\usepackage[final]{neurips_2021}

\usepackage[utf8]{inputenc} %
\usepackage[T1]{fontenc}    %
\usepackage[small]{caption}

\usepackage[dvipsnames,table]{xcolor}
\usepackage{natbib}
\usepackage{graphicx}
\usepackage{comment}
\usepackage{amsmath,amssymb} %
\usepackage{color}
\usepackage{bm}
\usepackage{xspace}
\usepackage{mathtools}
\usepackage{enumitem}
\usepackage{booktabs}
\usepackage{multirow}
\usepackage[bottom]{footmisc}
\usepackage{subcaption}
\usepackage{wrapfig}
\usepackage{soul}
\usepackage{amsthm}
\usepackage{nicefrac}       %
\usepackage{microtype}
\usepackage{amsbsy}
\usepackage{bbm}
\usepackage{stfloats}
\usepackage[linesnumbered,ruled,shortend]{algorithm2e}
\usepackage{mathrsfs}
\usepackage{thmtools}
\usepackage{thm-restate}
\usepackage{cleveref}
\usepackage{xr}
\usepackage{url}
\definecolor{Gray}{gray}{0.9}

\makeatletter
\newcommand{\removelatexerror}{\let\@latex@error\@gobble}
\makeatother

\makeatletter
\def\th@plain{%
  \thm@notefont{}%
  \itshape %
}
\def\th@definition{%
  \thm@notefont{}%
  \normalfont %
}
\makeatother

\newtheorem*{prop:early}{Proposition \ref{prop:early}}
\newtheorem*{prop:logistic}{Proposition \ref{prop:logistic}}

\newcommand{\norm}[1]{\left\lVert#1\right\rVert}
\DeclareMathOperator*{\argmin}{arg\,min}
\DeclareMathOperator*{\argmax}{arg\,max}

\DeclarePairedDelimiter{\inner}{\langle}{\rangle}
\DeclarePairedDelimiter\abs{\lvert}{\rvert}

\newcommand{\pq}{PQ\xspace}
\newcommand{\padv}{GeoAdEx\xspace}
\newcommand{\sit}{\citet{sitawarin20minnorm}\xspace}
\newcommand{\yang}{\citet{yang20knn}\xspace}
\newcommand{\wang}{\citet{wang19primal}\xspace}

\title{Adversarial Examples for $k$-Nearest Neighbor Classifiers Based on Higher-Order Voronoi Diagrams}

\author{
Chawin Sitawarin \\
UC Berkeley\\
\texttt{chawins@eecs.berkeley.edu} \\
\And Evgenios M. Kornaropoulos \\
George Mason University\\
\texttt{evgenios@gmu.edu} \\
\And Dawn Song \\
UC Berkeley\\
\texttt{dawnsong@cs.berkeley.edu} \\
\And David Wagner \\
UC Berkeley\\
\texttt{daw@cs.berkeley.edu} \\
}

\begin{document}

\maketitle

\begin{abstract}
  Adversarial examples are a widely studied phenomenon in machine learning models.
  While most of the attention has been focused on neural networks, other practical models also suffer from this issue.
  In this work, we propose an algorithm for evaluating the adversarial robustness of $k$-nearest neighbor classification, i.e., finding a minimum-norm adversarial example.
  Diverging from previous proposals, we propose the first geometric approach by performing a search that expands outwards from a given input point.
  On a high level, the search radius expands to the nearby higher-order Voronoi cells until we find a cell that classifies differently from the input point.
  To scale the algorithm to a large $k$, we introduce approximation steps that find perturbation with smaller norm, compared to the baselines, in a variety of datasets.
  Furthermore, we analyze the structural properties of a dataset where our approach outperforms the competition.
\end{abstract}

\section{Introduction}

It is well-known that machine learning models can easily be misled by maliciously crafted inputs, called adversarial examples, which are generated by adding a tiny perturbation to test samples~\citep{biggio13,szegedy13,goodfellow14explaining}. 
Adversarial examples are often studied in the context of neural networks, leaving the problem largely unexplored for other classifiers. 
The $k$-nearest neighbor, or simply $k$-NN, classifier is a simple yet widely used model in various applications such as data mining, recommendation, and anomaly detection systems where interpretability and simplicity are preferred~\citep{wu_top_2008}. 
This non-parametric classifier does not require a training phase and has a well-understood and elegant geometric foundation~\citep{10.5555/135734}. 
$k$-NN is also an active area of research with lots of developments in popular libraries like Google’s ScaNN~\citep{guo_accelerating_2020} and Facebook’s FAISS~\citep{johnson17faiss}. 
In recent works~\citep{papernot18dknn,dubey19webnn,sitawarin19knndef}, $k$-NN is combined with neural networks to enhance the robustness and the interpretability. 
Despite the importance of $k$-NN, there are only a handful of works that study its robustness~\citep{wang18knn,yang20knn,wang19primal,sitawarin19dknn,sitawarin20minnorm}.

The first step towards evaluating the robustness of a classifier is to generate an adversarial example that is close to a given test point, under some definition of closeness. 
In the case of $k$-NN where $k > 1$, this problem is challenging as the computation involves high-dimensional polytopes that satisfy a set of geometric properties, the so-called Voronoi cells. 

In this work, we propose the \padv algorithm for finding adversarial examples for $k$-NN classifiers. 
\padv is the first algorithm that exploits the underpinning geometry of the $k$-NN classifier. 
Specifically, our approach performs a principled search on the underlying high-order Voronoi diagram by expanding the search radius around the test point until it finds an adversarial, or simply incorrect, classification.
The geometric foundation of \padv allows us to locate nearby adversarial cells that the other attacks typically miss.
\padv follows the footsteps of the work by \citet{jordan19geocert} who exploited the geometry of neural networks for verifying the outputs. 
Our algorithmic approach stands in sharp contrast to previous attacks for $k$-NN~\citep{yang20knn,wang19primal} which refine the exhaustive approach by heuristic-based filtering.

Furthermore, we introduce optimizations and approximations to the main algorithm, and as a result, the experiments show that \padv discovers the smallest adversarial distance compared to all of the baselines in the vast majority of our experiments for $k \in \{3, 5, 7\}$. 
\padv finds up to $25\%$ smaller mean adversarial distance compared to the second best attack. 
We note that one inherent shortcoming of our geometric approach is the increased computation time, an aspect that can be further improved.

Finally, we present experiments that demonstrate that \padv performs significantly better when points from different classes are ``mixed" together, i.e., there is no clear separation between classes. 
On a high level, such a setup generates a more intricate spatial tessellation with nearby cells that alternate classes. 
In this challenging case, \padv performs up to $5\times$ better than the baselines.

\section{Background and Related Work} \label{sec:background}

\subsection{Background}

Let $X=(x_1,\ldots,x_n)$ be a set of points from $\mathbb{R}^d$, also called \emph{generators}, and let $Y=(y_1,\ldots,y_n)$ be the class of each point from a set of possible classes $\{1,\ldots,c\}$. 
Let $d(x,x')$ denote a metric between $x,x'\in\mathbb{R}^d$.
Let $Z^{(k)}(X)$ be the set of all possible subsets consisting of $k$ points from $X$, i.e., $Z^{(k)}(X)=\{L_1^{(k)},\ldots,L_i^{(k)},\ldots,L_l^{(k)}\}$, where $L_i^{(k)}=\{x_{i1},\ldots,x_{ik}\}$, $x_{ij}\in X$ and $l=\binom{n}{k}$. 
We define as \emph{order-k Voronoi cell} associated with $L_i^{(k)}$ the set $V(L_i^{(k)})$ that includes the points of $\mathbb{R}^d$ that are closer to $L_i^{(k)}$ than any other $k$ points of $X$. 
More formally:
\begin{align*}
V(L_i^{(k)})=\Big\{p \mid \max_{x}\{d(p,x)|x\in L_i^{(k)}  \} \leq  \min_{x'}\{d(p,x')|x'\in X \setminus L_i^{(k)}  \}  \Big\}
\end{align*}
With the term \emph{Voronoi facet}, or simply facet, we refer to a boundary of an order-$k$ Voronoi cell. 
We define as bisector $B\{x_a,x_b\}$ the set of points from $\mathbb{R}^d$ that are equidistant from $x_a\in X$ and $x_b\in X$. 
There is a tight connection between bisectors and facets. 
Every Voronoi facet is part of a bisector, but not every bisector, $\binom{n}{2}$ in total, includes a facet. 
We call the bisectors that do include a facet \emph{active} bisectors and those that do not \emph{inactive} bisectors. 
We also note that an order-$k$ Voronoi cell has at most $k(n-k)$ facets. 
The collection of order-k Voronoi cells for all $L_i^{(k)}$ is called \emph{order-k Voronoi diagram}. 
In this work, we assume the standard non-cocircularity assumption\footnote{In practice this can be achieved by adding a small random noise on each generator to break potential cocircularity between them.} to avoid degenerate cases. 
We only consider the Euclidean distance for the $k$-NN, i.e., $d(x_a, x_b) = \|x_a - x_b\|_2$.

Up to this point, we have not used the classes of the $X$. 
In this work, we focus on $k$-NN classifiers that classify based on the majority (class) vote of the $k$-nearest points. 
We define as \emph{adversarial cell} with respect to $(x,y)$ any order-$k$ Voronoi cell that has different majority than label $y$. 
We denote with $A(x)$ the set of all adversarial cells with respect to $(x,y)$. 
With the term \emph{adversarial facet} with respect to $(x,y)$ we refer to a facet that belongs to a cell from $A(x)$.

\subsection{Related Work}
\paragraph{Adversarial Robustness of $\bm{k}$-NN Classifiers.} \citet{wang18knn} study the robustness property of $k$-NN classifiers from a theoretical perspective. The authors show that a $k$-NN classifier can be as robust as the optimal Bayes classifier given a sufficiently large number of generators and $k$. \citet{wang18knn} and \citet{yang20knn} also propose potential methods for improving the robustness of $k$-NN by pruning away some of the ``ambiguous'' generators. Other works consider different alternatives for enhancing the robustness of neural networks by combining them with $k$-NN. \citep{papernot18dknn,sitawarin19knndef,dubey19webnn}

\paragraph{Attacks on $\bm{k}$-NN Classifiers.} \citet{sitawarin19dknn,sitawarin20minnorm} propose a method for finding minimum-norm adversarial examples on $k$-NN and deep $k$-NN classifiers~\citep{papernot18dknn}. Their method approximates a $k$-NN classifier with a soft differentiable function so adversarial examples can be found via a gradient-based optimization algorithm. While the approach is efficient and scalable to large $k$, it provides no guarantee for finding the nearest adversarial example and overlooks the geometry of $k$-NN.

\citet{yang20knn} and \citet{wang19primal} take a similar approach for finding the adversarial example closest to a given test $x$. Their method follows an exhaustive search by computing the distance between $x$ and all the Voronoi cells that have a different class from $x$. Both of them show that this can be done exactly when $k=1$ but does not scale well for $k > 1$. Consequently, they introduce heuristics to improve the efficiency by choosing only a subset of the Voronoi cells. However, in the process, they lose the optimality guarantee on the resulted adversarial example. More importantly, the heuristics will miss an exponential number of cells as $k$ increases. 

\paragraph{Attacks Based on Geometric Insights.}
The problem of verifying or finding the nearest adversarial examples in neural networks is an active research direction where solutions still do not scale to a satisfactory degree \citep{huang17verify,katz17reluplex,tjeng17mip,wong18polytope,weng18cert,cohen19smooth,jordan19geocert}. 
Among these works, \citet{jordan19geocert} proposed an alternative for a ReLU network, using a geometric approach. 
Specifically, the authors notice that piecewise linear networks partition the input space into numerous polytopes where points of the same polytope are classified with the same label. 
Armed with this observation, they propose an algorithm that iteratively searches through these polytopes that are further away from the test sample until they find a polytope with a different label.

\section{The \padv Algorithm}

\paragraph{Objective.} The goal of the algorithm is to find the smallest perturbation $\delta^*$ that moves a test point $(x,y)$ to an adversarial cell. 
This objective can be expressed as the following optimization problem:
\begin{align} \label{eq:main_opt}
    \delta^* ~=~ \argmin_{\delta}& \quad \norm{\delta}_2^2 \\
    \text{s.t.}& \quad x + \delta \in A(x) \nonumber
\end{align}
We define as $\epsilon^*$ \emph{optimal adversarial distance} where $\epsilon^* \coloneqq \norm{\delta^*}_2$. 
The norm of any other (non-optimal) perturbation $\delta$ that misclassifies $(x,y)$, i.e., $x + \delta \in A(x)$, is simply called \emph{adversarial distance}.

\begin{figure}[t]
    \centering
    \includegraphics[width=0.5\textwidth]{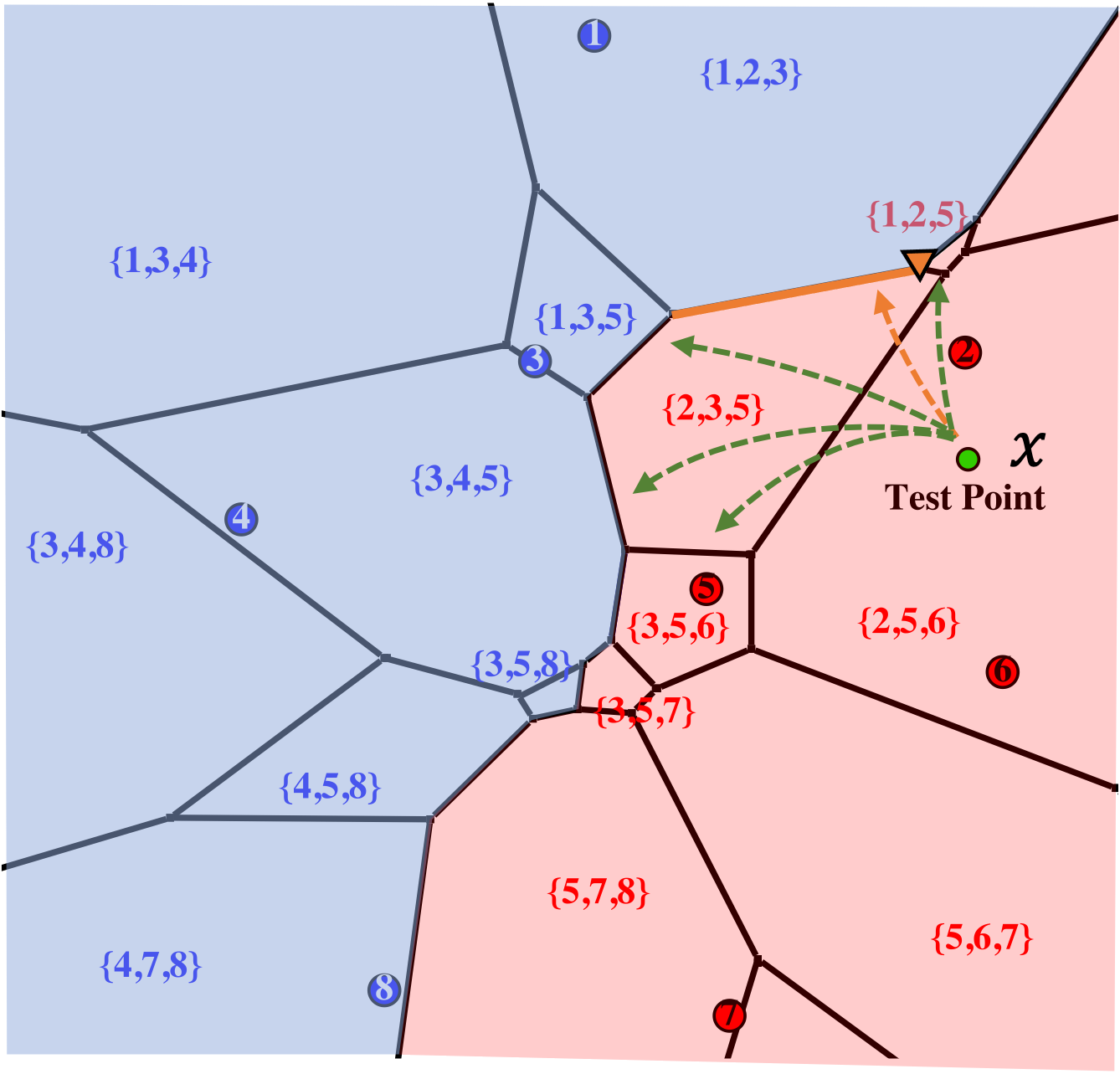}
    \caption{\padv on a Voronoi diagram for $k=3$ in $\mathbb{R}^2$ with two classes. The color of each numbered generator point shows its class, and the color of each cell indicates its classification outcome. The illustration presents the step where \padv has processed the facets of $V(\{2,5,6\})$ and is transitioning to the next closest cell $V(\{2,3,5\})$. The arrows indicate the facets that are inserted to the priority queue $\pq$. When \padv terminates, it outputs the closest adversarial facet (orange line); this transition is indicated with the orange arrow, and the orange triangle is the optimal adversarial example for $x$.}
    \vspace{-5pt}
    \label{fig:main_diagram}
\end{figure}

\paragraph{A First Approach.} The constraint of the above formulation implies that $x+\delta$ must be a member of an adversarial cell from $A(x)$. 
A simple approach to solve this minimization is to build a series of optimization problems, one for each of the cells in $A(x)$, and pick the solution with the minimum adversarial distance. 
Unfortunately, this would require solving $O(\binom{n}{k})$ quadratic programs each of which has $k(n-k)$ constraints. 
While this complexity may be manageable when $k = 1$ and $n$ is small, it does not scale with $k$. 
On a high level, \yang and \wang take this approach and additionally develop heuristics to scale to cases with $k > 1$.

\paragraph{The \padv Approach.}
The core idea of \padv is to perform a principled \emph{geometric exploration} around the test point $x$. 
\padv processes order-$k$ Voronoi cells \`a la breadth-first search until it discovers an adversarial cell. 
Algorithm~\ref{algo:padv} provides a pseudocode of the main steps of \padv, and Figure~\ref{fig:main_diagram} illustrates some steps of \padv on a $3$-NN classifier with two classes.
In the following subsections, we explain these steps in detail and describe  performance speedups as well as  approximation steps to scale the algorithm to $k > 1$.

\begin{algorithm}[ht]
    \small
 	\caption{\label{algo:padv} 	\padv Algorithm }
	\KwData{Test point $(x,y)$}
 	\KwResult{Adversarial distance $\epsilon$}
 	Initialize the smallest adversarial distance $\epsilon$ found so far as $\epsilon \leftarrow \infty$\;
     Initialize $\pq$ by inserting the facets of the order-$k$ Voronoi cell that $x$ falls into. Mark this cell as \emph{visited}\;
	Call \emph{deleteMin} from $\pq$ until the returned facet is part of an unvisited order-$k$ Voronoi cell. Initialize $\Psi$ to this unvisited Voronoi cell\label{line:PsiDiscovery}\; 
	\While{$\Psi$ is not an adversarial cell}{
	Find the set of facets $\Phi$ that comprise unvisited cell $\Psi$ (Section~\ref{ssec:find_neighbors})\label{line:Facets}\;
	Compute the distance between query point $x$ and each of the facets in $\Phi$ (Section~\ref{ssec:distance_compute})\;
	If (i) a facet from $\Phi$ is an adversarial facet, and (ii) the distance to this adversarial facet implies a smaller norm than $\epsilon$, then update $\epsilon$\;
    Insert to $\pq$ the facets in $\Phi$ with their distance if smaller than $\epsilon$\;
    	Call \emph{deleteMin} from $\pq$ until the returned facet is part of an unvisited order-$k$ Voronoi cell. Update $\Psi$ to the new unvisited Voronoi cell\label{line:PsiDiscovery2}\; 
	}
	If an adversarial facet is removed from $\pq$ via \emph{deleteMin}, the algorithm return the optimal adversarial distance (see Lemma~\ref{thm:correct})\;
	\Return{$\epsilon$}\;
\end{algorithm}

\subsection{Data Structures}
We maintain a priority queue $\pq$ as an auxiliary data structure that is defined for a given test point $(x,y)$. $\pq$ tracks the progress made so far towards locating the optimal adversarial distance. 
Specifically, $\pq$ performs the operations \emph{insert} (resp. \emph{deleteMin}) where the input (resp. output)  is a facet of the order-$k$ Voronoi diagram of $X$. 
Every facet in $\pq$ is accompanied by the distance between the test point $x$ and the facet.
Priority is given to the facet with the \emph{minimum} distance to $x$. 
We also need to mark which Voronoi cells are processed. 
For that, we use a hash table (Python's native dictionary object) that has an entry for every order-$k$ Voronoi cell that the algorithm has visited. 
We refer to Voronoi cells that are part of the dictionary as \emph{visited} and those that are not as \emph{unvisited}.

\subsection{Find Neighbors of a Voronoi Cell} \label{ssec:find_neighbors}

Line~\ref{line:PsiDiscovery} \&~\ref{line:PsiDiscovery2} of Algorithm~\ref{algo:padv}, \padv discovers an \emph{unvisited} order-$k$ Voronoi cell $\Psi$. 
In Line~\ref{line:Facets}, \padv needs to find the facets that comprise $\Psi$, which is equivalent to finding the Voronoi cells that are \emph{neighboring} with cell $\Psi$. 
We present three options for implementing the discovery of neighbors; the choosing among them depends on the application, the size of the dataset, and the value of $k$.

\paragraph{(A) Neighbors via Order-$\bm{k}$ Voronoi Diagram.} 
The exact computation of an order-$k$ Voronoi diagram in high dimensions constitutes a computationally challenging task. 
The cost of this option amortizes when there is a large number of test points to be processed, and the dataset is fairly small in size and dimension. However, it is highly unlikely to scale when $k > 1$ which is the goal of our work.

\paragraph{(B) Neighbors via Enumerating Bisectors.}
In this approach, one has to process a quadratic number of bisectors that are potentially \emph{active} with respect to the Voronoi cell $\Psi$ (see the definition from Section~\ref{sec:background}). 
Interestingly, we can filter out this set by only considering the bisectors whose generator, exactly one of the two, also generates cell $\Psi$.
Consequently, the task of finding the facets of the current cell $\Psi$ is reduced to testing the \emph{activeness} of $k(n-k)$ bisectors. 
Jumping ahead, this step can be accomplished simultaneously with computing distance from $x$ to each bisector so we defer its analysis for Section~\ref{ssec:distance_compute}. 
For comparison, the complexity of this step is $O(poly(n,d,k))$ without any speedups, and in the worst case, which is extremely unlikely in practice, it can be called up to $N$ times where $N$ is the number of all cells ($N$ is $O(\binom{n}{k})$).

\paragraph{(C) Neighbors via Order-1 Voronoi Diagram.} 
This approach is the middle ground between the first two. 
That is, it still tests whether a bisector is active, much like (B), but it uses the Voronoi diagram, much like (A), to reduce the number of bisectors to test. 
Given a Voronoi cell with $k$ generators, e.g.,  $V(\{x_1,\dots,x_{k-1},x_k\})$, we know that its neighboring cells must have a set of generators that differs by exactly one point, e.g., $V(\{x_1,\dots,x_{k-1},x_l\})$. We draw a connection between the Voronoi diagrams of order 1 and $k$ and show that the new point $x_l$ \emph{must be neighboring with at least one of the order-1 cells} $V(\{x_1\}), V(\{x_2\}), \ldots, V(\{x_k\})$. 
We formalize this in Theorem~\ref{thm:firstorder}, and its proof can be found in Appendix~\ref{ap:proof}.

\begin{restatable}{thm}{firstorder}
\label{thm:firstorder}
Let $S=\{x_1,\ldots,x_{k-1}\}\subset X$ be a set of $k-1$ generators. Let $x_k,x_l\in X$ be two generators such that $x_k,x_l\notin S$. 
If $V(S\cup \{x_k\})$ and $V(S\cup \{x_l\})$ are two neighboring order-$k$ Voronoi cells, 
then the order-$1$ Voronoi cell $V(\{x_l\})$ is neighboring with at least one of the $V(\{x_1\}),\ldots,V(\{x_{k-1}\}),V(\{x_k\})$.
\end{restatable}

With this insight, we can \emph{narrow down significantly} the number of bisectors considered in approach~(B).
More formally, let $nb(x')$ denote the set of generators of order-1 Voronoi cells that neighbor with $V\{x'\}$. 
Let $\Psi$ be the Voronoi cell $V(\{x_1,\ldots,x_k\})$. 
By Theorem~\ref{thm:firstorder}, it is enough to consider the bisectors between $z \in \{x_1,\dots,x_k\}$ and $z' \in \left(  \bigcup_{i=1}^k nb(x_i)\right)-\{x_1,\dots,x_k\}$. 
Hence, the total number of bisectors we need to test is at most $k \cdot \abs{\bigcup_{i=1}^k nb(x_i)}$.
Assuming that $\ell \coloneqq \max_i \abs{nb(x_i)}$, the number of bisectors cannot exceed $k^2\ell$. 
Hence, approach~(C) is much faster than approach~(B) when $k^2\ell \ll n$. 
 In this work, we use approach~(B) for the $k=1$ case. However, in order to scale to large datasets and $k > 1$, we propose some approximations steps that are rooted in approach~(C) later in Section~\ref{sssec:approx}.

\subsection{Computing Distance \& Testing Activeness} \label{ssec:distance_compute}

Now we describe Line~6 of Algorithm~\ref{algo:padv}, i.e. how to compute the distance between a point $x$ and a facet. 
Then, expanding on this process, we detail how to test the activeness of bisectors.

\paragraph{Distance computation.} 
A Voronoi cell is a polytope that can also be described as an intersection of halfspaces: $\{z \mid Az \le b\}$.
A bisector can be described as a hyperplane: $\{z \mid\inner{z, \hat{a}} = \hat{b}\}$. 
The corresponding facet needs to both be part of the polytope \emph{and} satisfy the equation of the hyperplane. Thus, the shortest distance from $x$ to a facet is given by the square root of the optimum of the following quadratic program:
\begin{align} \label{eq:distance}
    \min_{z}& \quad \norm{z - x}_2^2 \\
    \text{s.t.}& \quad Az \le b ~\text{and}~ \inner{z, \hat{a}} = \hat{b} \nonumber
\end{align}
This problem can be solved by any off-the-shelf solver, but in Appendix~\ref{ap:dual}, we describe a faster method also used by \wang.

\paragraph{Testing Bisector Activeness.} 
Assuming that $\{z \mid Az \le b\} \ne \emptyset $, the bisector $\inner{z, \hat{a}} = \hat{b}$ is active \emph{if and only if} Eqn.~\eqref{eq:distance} is feasible since their intersection has to be non-empty.
In case this problem is infeasible, the bisector is inactive. 
From duality theory, we know that Eqn.~\eqref{eq:distance}, i.e. the primal, is feasible if and only if the dual objective is unbounded since the dual problem is always feasible in this case.
Checking the unboundedness can be accomplished as we are solving Eqn.~\eqref{eq:distance} in its dual form so we can combine the steps of ``distance computation" and ``testing activeness" into a single function that either returns the distance or indicates that the bisector is inactive.

\subsection{Optimality of \padv} \label{ssec:optimality}

Here, we formally state two lemmas that together prove the output optimality of \padv.
The next lemma is straightforward  as it is a direct consequence of Lemma C.2 from \citet{jordan19geocert}. 
For completion, we prove it in Appendix~\ref{ap:proof}.

\begin{restatable}[Correctness of \padv]{lemma}{correct}
\label{thm:correct}
Provided no time limit, \padv terminates when it finds the optimal adversarial examples or equivalently, one of the solutions of Eqn.~\eqref{eq:main_opt}.
\end{restatable}

The next lemma states that if \padv terminates early, i.e. in case we enforce a time limit for performance purposes (see Section~\ref{sssec:approx}), it can still guarantee a lower bound to $\epsilon^*$.
Specifically, the distance to $x$ from the most recent facet deleted from $\pq$ serves as a lower bound to $\epsilon^*$. 

\begin{restatable}[Lower bound guarantee]{lemma}{lowerbound}
\label{thm:lowerbound}
If \padv terminates early, the distance from test point $x$ to the last deleted facet from $\pq$ is a lower bound to the optimal adversarial distance $\epsilon^*$.
\end{restatable}

We note that if \padv terminates early, then $\epsilon$ serves as an upper bound of $\epsilon^*$. 
None of the previous approaches on adversarial examples for $k$-NN provide both an upper and lower bound guarantee.

\subsection{Towards Scaling \padv} \label{ssec:scale}

For simplicity, in Algorithm~\ref{algo:padv}, we present the main algorithmic steps that need to be performed to compute the optimal adversarial distance using the geometric structure of the problem. 
In this section we delve into details on how to accelerate the performance of \padv by \emph{performance optimization} and \emph{approximation} steps.

\subsubsection{Optimizing Computation Time} \label{sssec:time_optimize}

We introduce four performance optimizations that significantly speed up \padv. We present the two most important ones here and the rest in Appendix~\ref{ap:opt}.

\paragraph{(I) Pruning Distant Facets.} Given \emph{any} adversarial distance, e.g., the intermediate result $\epsilon$ from Algorithm~\ref{algo:padv}, we know  that any facet that is more than $\epsilon$ afar from $x$ is not the facet associated with the optimal adversarial distance, i.e., $\epsilon$ acts as an upper bound that is refined during the execution. 
Hence, we can safely filter out these facets, and it is unnecessary to compute their distance to $x$, test activeness, or insert them to $\pq$. 
This technique can be used both before and during the facet distance computation (Section~\ref{ssec:distance_compute}).
A benefit of our principled geometric approach is that we can apply geometric arguments to eliminate redundant computation on Voronoi cells that are far away.

\paragraph{(II) Rethinking the Initialization of $\bm{\epsilon}$.} 
Recall that Line~1 of Algorithm~\ref{algo:padv} initializes $\epsilon$ to $\infty$. 
Given the upgraded role of $\epsilon$ in the previous paragraph, it is clear that a non-simplistic initialization would filter out more unnecessary computation early on and scale the overall performance.
For our experiments, we run \sit to initialize $\epsilon$ since it yields a reasonable $\epsilon$ and is faster than the other attacks.

\subsubsection{Acceleration via Approximations} \label{sssec:approx}

As mentioned in Section~\ref{ssec:find_neighbors}, approach~(C) for finding the neighbors of an order-$k$ Voronoi cell still requires the knowledge of the first-order cells. 
This can be obtained by either computing the entire order-1 Voronoi diagram or enumerating all possible facets of order-1 cells. 
Nonetheless, first-order cells in a high-dimension dataset may have a large number of neighbors, and building a Voronoi diagram can easily become a bottleneck in high-dimensions. 
In this case, approach~(C) is no better than approach~(B) of Section~\ref{ssec:find_neighbors}. To scale to large and high-dimension datasets, we introduce some approximations. Here, we propose the \emph{approximate version} of \padv built upon the relationship between order-1 and order-$k$ neighbors from Theorem~\ref{thm:firstorder}. 

\paragraph{Description.} To circumvent the expensive (in high dimensions) computation of the order-1 neighbors, we approximate approach~(C) of Section~\ref{ssec:find_neighbors}. 
Specifically, instead of operating on the neighbors of the order-1 cell $V\{x_i\}$, we operate on a subset of $m$ points chosen from the entire set of $X$ according to a fast heuristic. 
In other words, $nb(x_i)$ is (roughly) approximated by a new subset of fixed size $m$ denoted as $\widetilde{nb}_m(x_i)$.
For cell $V\{x_1,\dots,x_k\}$ and a given generator $x_i$ of this cell, we select from $X$ the $m$ closest points to $x_i$ that are not in $\{x_1,\dots,x_k\}$. Mathematically, we define $\widetilde{nb}_m(x_i)$ as:
\begin{align*}
    \widetilde{nb}_m(x_i) = \left\{x_j \in X \mid d(x_i, x_j) \le  d(x_i, x_{\pi(m)}) \right\} \setminus \{x_1,\ldots,x_k\}
\end{align*}
where $x_{\pi(m)}$ is the $m$-th nearest neighbor of $x_i$.
With the above heuristic we guarantee that for each cell, we only have to compute the distance (and test activeness) for at most $k^2m$ facets and, thus, sidestep the computation of the first order Voronoi diagram in high dimensions. 
Each subsequent optimization problem has at most $k^2m$ constraints.
Finding $m$ nearest points to a single point $x_i$ is a well studied problem and can be \emph{approximately} solved very fast~\citep{johnson17faiss,aumuller17ann,andoni18approximate}.

\paragraph{Limitations due to Approximation.} 
Due to the above approximation, it is possible that some of the true order-1 neighbors are not included in $\bigcup_{i=1}^k \widetilde{nb}_m(x_i)$. 
Hence, some active facets and cells may be missed completely\footnote{Depending on the dataset and $m$, the chance of this happening may not be high for two reasons: (i) For a large $k$, $\bigcup_{i=1}^k \widetilde{nb}_m(x_i)$ becomes a large set and is likely to cover most, if not entire, $\bigcup_{i=1}^k nb(x_i)$. (ii) Even if we miss some cells as neighbors of a particular order-$k$ cell, they may still be picked up by the other cells.}. 
This leads to two limitations. The first limitation is that we can no longer guarantee the optimality of \padv through Lemmas~\ref{thm:correct} and \ref{thm:lowerbound}. In other words, we cannot conclude whether the adversarial distance returned by the approximate version of \padv is the optimal adversarial distance (or its lower bound). 

The second limitation is that the approximation may affect the correctness of the distance computation. 
Recall that in the exact version of \padv, we have shown that it is possible to replace the bisectors in the constraint of the optimization problem in Eqn.~\eqref{eq:distance} with bisectors between $\{x_1,\dots,x_k\}$ and $\bigcup_{i=1}^k nb(x_i)$.
But with the approximation, the new feasible set in Eqn.~\eqref{eq:distance} becomes a \emph{superset} of the one in the exact version. Consequently, the steps in Section~\ref{ssec:distance_compute} may falsely label an inactive bisector as active or return smaller distance than the true value.

\paragraph{Addressing Limitations.} The first limitation is inherent to the deployment of heuristics to boost performance; a similar issue appears in \sit as well as the approximate version of \yang and \wang.
It is difficult to avoid without increasing the runtime significantly. 
However, the second limitation can be addressed, and we do so by using the full set of bisectors when computing the distance and testing the activeness of adversarial facets like in the exact version. 
This incurs only a small computational cost.
We do not need to do the same for non-adversarial cells since their distance and activeness do not affect the correctness of \padv.
\begin{table}[!t]
\centering
\caption{Mean norm of the adversarial perturbations on 100 random test points across datasets (lower is better). We report the numbers averaged over 10 runs with random training and test splits. The error, highlighted in gray, is the 95\%-confidence interval. The numbers in parentheses is the ratio of the mean perturbation norm found by \padv over that of the best baseline. The smallest mean perturbation norm among the attacks for each dataset and each $k$ is bolded.}
\vspace{5pt}
\label{tab:main}
\small
\begin{tabular}{@{}llrrrrrrr@{}}
\toprule
$k$                & Attacks          & Australian & Covtype      & Diabetes & Fourclass & Gaussian & Letters & fmnist06 \\ \midrule
& S\&W & $.4242$ & $.1931$ & $.1055$ & $.1079$ & $.0442$ & $.1128$ & $.1554$ \\
\rowcolor{Gray}
\cellcolor{white} & \cellcolor{white} & $\pm .0201$ & $\pm .0148$ & $\pm .0068$ & $\pm .0039$ & $\pm .0028$ & $\pm .0049$ & $\pm .0121$ \\
& Yang et al. & $.4658$ & $.2385$ & $.1392$ & $.1209$ & $.1138$ & $.1370$ & $.1773$ \\
\rowcolor{Gray}
\cellcolor{white} & \cellcolor{white} & $\pm .0258$ & $\pm .0101$ & $\pm .0077$ & $\pm .0055$ & $\pm .0023$ & $\pm .0046$ & $\pm .0043$ \\
& Wang et al. & $.4466$ & $.2134$ & $.1164$ & $.1117$ & $.0825$ & $.1218$ & $.1643$ \\
\rowcolor{Gray}
\cellcolor{white} & \cellcolor{white} & $\pm .0205$ & $\pm .0170$ & $\pm .0052$ & $\pm .0039$ & $\pm .0030$ & $\pm .0050$ & $\pm .0047$ \\
& \textbf{\padv} & $\bm{.3646}$ & $\bm{.1448}$ & $\bm{.0786}$ & $\bm{.1073}$ & $\bm{.0426}$ & $\bm{.1091}$ & $\bm{.1509}$ \\
\rowcolor{Gray}
\cellcolor{white} & \cellcolor{white} & $\pm .0250$ & $\pm .0116$ & $\pm .0042$ & $\pm .0045$ & $\pm .0015$ & $\pm .0066$ & $\pm .0076$ \\
\multirow{-9}{*}{3} & & \textcolor{ForestGreen}{(.8595)} & \textcolor{ForestGreen}{(.7499)} & \textcolor{ForestGreen}{(.7450)} & \textcolor{ForestGreen}{(.9944)} & \textcolor{ForestGreen}{(.9638)} & \textcolor{ForestGreen}{(.9672)} & \textcolor{ForestGreen}{(.9710)} \\ \cmidrule(l){2-9}
& S\&W & $.4748$ & $.2281$ & $.1215$ & $.1087$ & $.0463$ & $.1134$ & $.1648$ \\
\rowcolor{Gray}
\cellcolor{white} & \cellcolor{white} & $\pm .0199$ & $\pm .0121$ & $\pm .0060$ & $\pm .0052$ & $\pm .0030$ & $\pm .0064$ & $\pm .0093$ \\
& Yang et al. & $.5524$ & $.3047$ & $.1824$ & $.1309$ & $.1776$ & $.1503$ & $.2087$ \\
\rowcolor{Gray}
\cellcolor{white} & \cellcolor{white} & $\pm .0141$ & $\pm .0161$ & $\pm .0068$ & $\pm .0043$ & $\pm .0031$ & $\pm .0048$ & $\pm .0059$ \\
& Wang et al. & .$.5110$ & $.2613$ & $.1382$ & $.1127$ & $.1195$ & $.1298$ & $.1877$ \\
\rowcolor{Gray}
\cellcolor{white} & \cellcolor{white} & $\pm .0147$ & $\pm .0110$ & $\pm .0056$ & $\pm .0054$ & $\pm .0047$ & $\pm .0055$ & $\pm .0088$ \\
& \textbf{\padv} & $\bm{.4608}$ & $\bm{.1856}$ & $\bm{.1021}$ & $\bm{.1066}$ & $\bm{.0401}$ & $\bm{.1130}$ & $\bm{.1632}$ \\
\rowcolor{Gray}
\cellcolor{white} & \cellcolor{white} & $\pm .0175$ & $\pm .0218$ & $\pm .0065$ & $\pm .0048$ & $\pm .0023$ & $\pm .0021$ & $\pm .0075$ \\
\multirow{-9}{*}{5} & & \textcolor{ForestGreen}{(.9705)} & \textcolor{ForestGreen}{(.8137)} & \textcolor{ForestGreen}{(.8403)} & \textcolor{ForestGreen}{(.9981)} & \textcolor{ForestGreen}{(.8661)} & \textcolor{ForestGreen}{(.9965)} & \textcolor{ForestGreen}{(.9903)} \\
\cmidrule(l){2-9} 
& S\&W & $\bm{.5110}$ & $.2528$ & $.1259$ & $.1129$ & $.0463$ & $\bm{.1127}$ & $\bm{.1662}$ \\
\rowcolor{Gray}
\cellcolor{white} & \cellcolor{white} & $\pm .0145$ & $\pm .0211$ & $\pm .0089$ & $\pm .0043$ & $\pm .0022$ & $\pm .0049$ & $\pm .0051$ \\
& Yang et al. & $.6010$ & $.3335$ & $.2005$ & $.1403$ & $.2204$ & $.1637$ & $.2328$ \\
\rowcolor{Gray}
\cellcolor{white} & \cellcolor{white} & $\pm .0140$ & $\pm .0245$ & $\pm .0097$ & $\pm .0053$ & $\pm .0021$ & $\pm .0035$ & $\pm .0057$ \\
& Wang et al. & .$.5410$ & $.3078$ & $.1631$ & $.1182$ & $.1570$ & $.1395$ & $.2010$ \\
\rowcolor{Gray}
\cellcolor{white} & \cellcolor{white} & $\pm .0181$ & $\pm .0145$ & $\pm .0083$ & $\pm .0051$ & $\pm .0017$ & $\pm .0034$ & $\pm .0056$ \\
& \textbf{\padv} & $.5134$ & $\bm{.2153}$ & $\bm{.1202}$ & $\bm{.1109}$ & $\bm{.0425}$ & $.1151$ & $.1692$ \\
\rowcolor{Gray}
\cellcolor{white} & \cellcolor{white} & $\pm .0149$ & $\pm .0174$ & $\pm .0049$ & $\pm .0019$ & $\pm .0033$ & $\pm .0065$ & $\pm .0055$ \\
\multirow{-9}{*}{7} & & \textcolor{red}{(1.0047)} & \textcolor{ForestGreen}{(.8517)} & \textcolor{ForestGreen}{(.9547)} & \textcolor{ForestGreen}{(.9822)} & \textcolor{ForestGreen}{(.9081)} & \textcolor{red}{(1.0213)} & \textcolor{red}{(1.0181)} \\
\bottomrule
\end{tabular}
\end{table}

\section{Experiments} \label{sec:experiment}

We compare \padv to all of the previously proposed attacks on $k$-NN classifiers, namely \sit, \yang, and \wang. 
The attacks are evaluated on seven datasets most of which were used in the experiments by \yang. 
Namely, we evaluate the attacks in that following datasets: Australian, Covtype, Diabetes, Fourclass, and fmnist06 (a two-class subset, `0' and `6', of Fashion-MNIST). 
Additionally, we also evaluate on Gaussian and Letters datasets which have more classes and data points.
More details on the datasets and the computational setup are included in Appendix~\ref{ap:exp_detail}.

\subsection{Main Findings}

Table~\ref{tab:main} compares the proposed \padv attack against the three baselines with respect to the mean perturbation norm. 
Interestingly, \padv \emph{outperforms the other baselines} on all of the seven datasets for $k=3,5$ and on four of them for $k=7$. 
In the three remaining datasets, \sit performs best and is narrowly better than \padv. 
\padv performs notably well on datasets Gaussian and Covtype with up to $25\%$ smaller perturbation norm than the second best attack. 
On average, \padv reduces the perturbation norm by $11\%, 8\%,$ and $4\%$ for $k=3,5,7$, respectively. 
Since the results of the exact attacks for $k=1$ are not the main focus of this work, they appear in Appendix~\ref{app:k1}.

We report that while our attack finds an adversarial distance that is closer to the optimal compared to the baselines, its main limitation is the runtime. 
In some cases, the runtime of \padv can be an order of magnitude larger than the second best attack. 
This is discussed in more detail in Section~\ref{ssec:runtime}.

It is also important to note that each of the proposed attacks (including \padv) performs better when the underlying data present certain structural properties. 
As a result, there is no single attack, so far, that performs universally better across all datasets. 
In Section~\ref{ssec:structure}, we explore the property of datasets that make our attack superior.

\subsection{Advantages of a Geometric Search} \label{ssec:structure}

Intuitively, \padv is based on a search that expands outwards from the original test point. 
As a result, it performs significantly better than the baselines when there exists an adversarial cell in relatively close proximity to the given test point. 
In other words, \padv performs really well on datasets when the class-conditioned distributions are closer or present significant ``overlap.''
In this case, \padv is less likely to miss small adversarial cells that are close to the test point as confirmed by the main experiment. 
We verify this hypothesis in the following experiment where we control the closeness of classes and study the performance of \padv compared to the baselines.

\paragraph{Experiments with Class Closeness.} 
Given a dataset with $c$ class-conditioned data distributions $D_1,\dots,D_c$, we can compute the minimum KL-divergence between a 
pair of distributions $D_i$ and $D_j$
: $\textsf{KLD}(D_i||D_j)$. Now we define as \emph{class closeness} of a given data distribution as the average of the above set of minima, i.e., $\frac{1}{c}\sum_{i=1}^c\min_j \textsf{KLD}(D_i||D_j)$. 
The smaller the class closeness, the closer their distributions are and, consequently the larger the degree of ``mixing'' between Voronoi cells of different classes. 
For each dataset, we record the associated class closeness and the ratio of the mean perturbation norm of each baseline over that of \padv.

\begin{figure}[t]
    \centering
    \hfill
    \begin{subfigure}[t!]{0.42\textwidth}
        \centering
        \includegraphics[width=\textwidth]{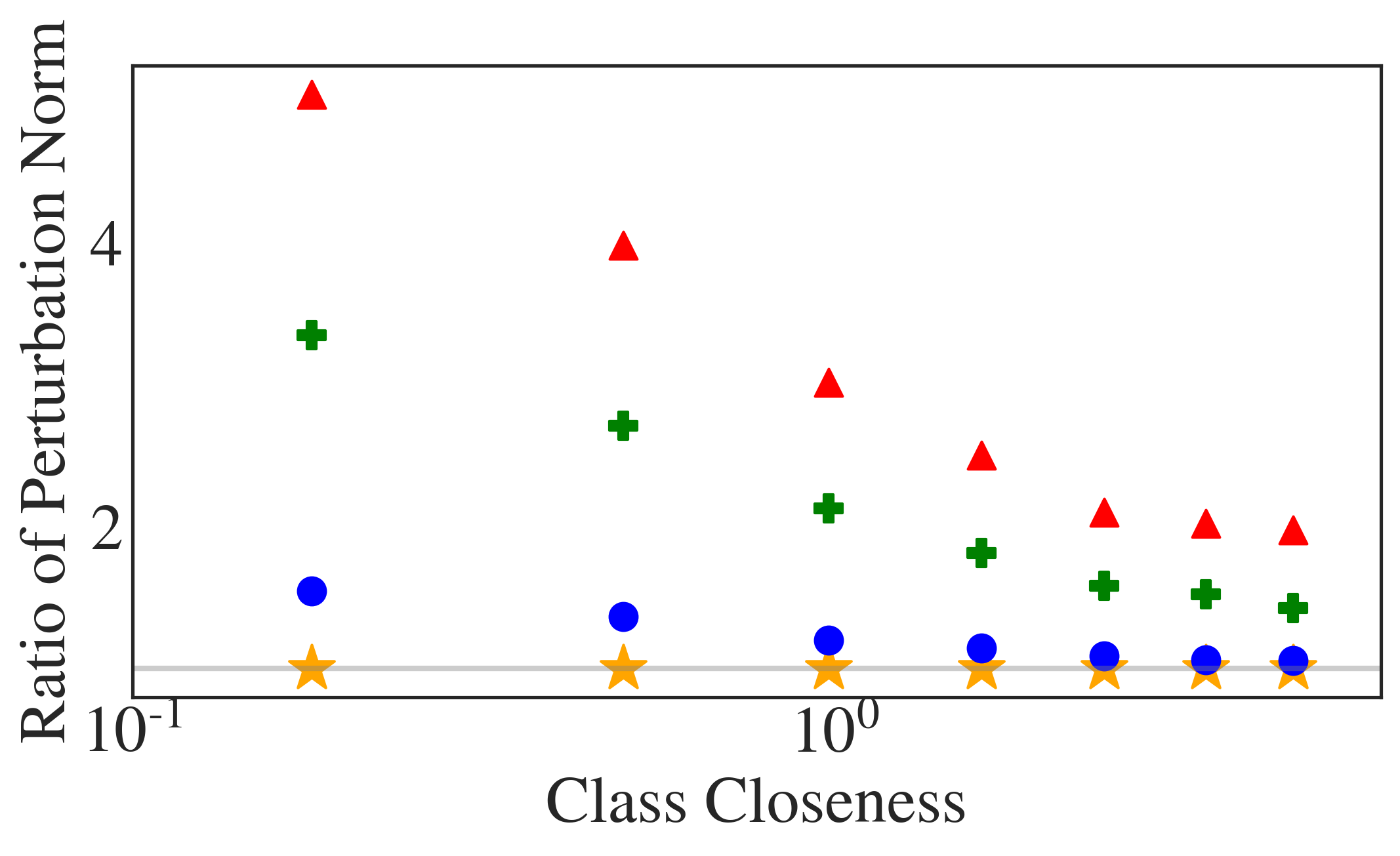}
        \vspace{0.34cm}
        \caption{Gaussian ($k=5$)}
        \label{fig:kld_gauss}
    \end{subfigure}
    \hfill
    \begin{subfigure}[t!]{0.42\textwidth}
        \centering
        \includegraphics[width=\textwidth]{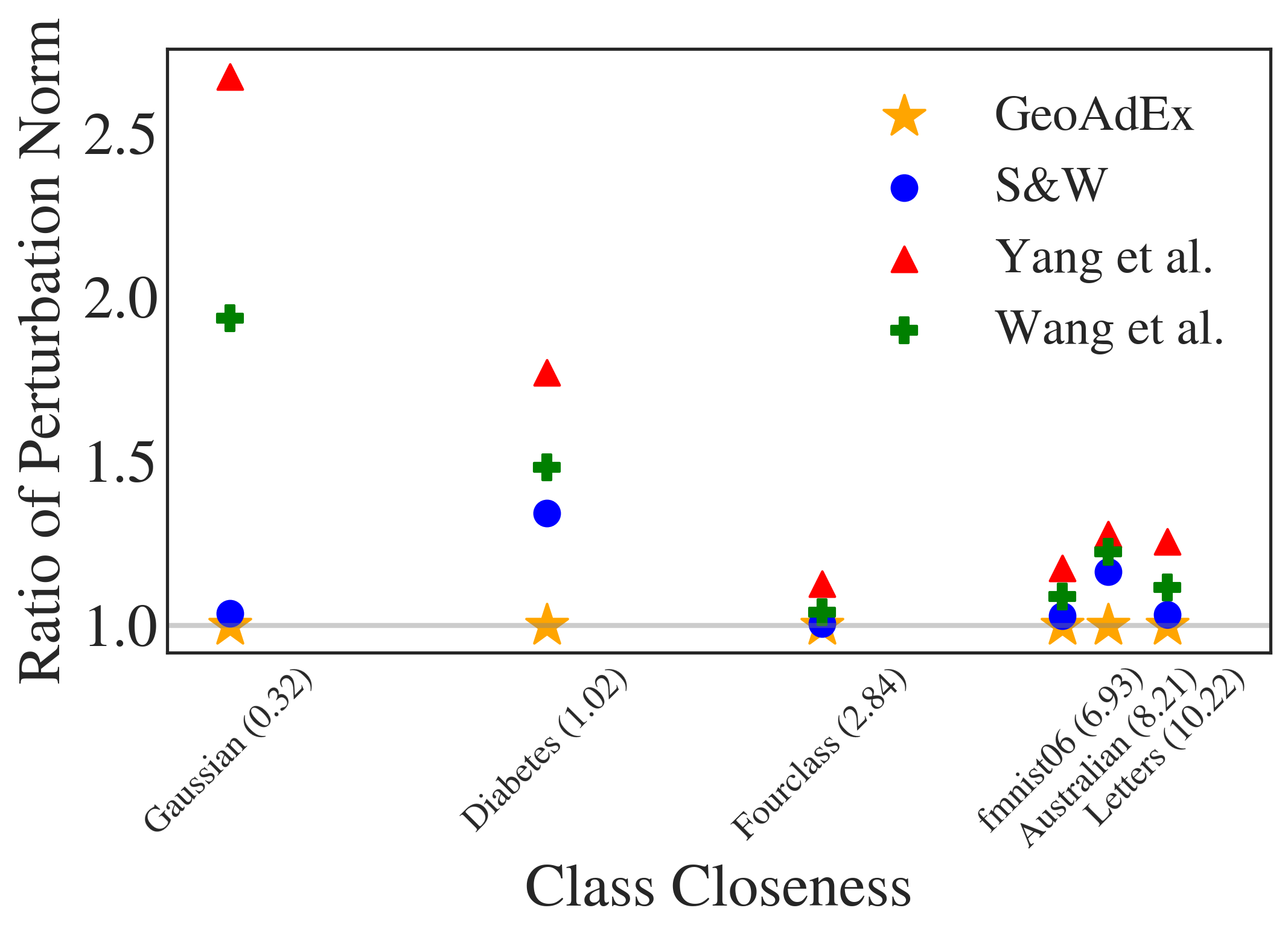}
        \caption{Real datasets ($k=3$)}
        \label{fig:kld_others}
    \end{subfigure}
    \hfill~
    \caption{Mean perturbation norm found by the attacks as a function of the class closeness (lower is better).}
    \label{fig:kld}
\end{figure}

First, we confirm this hypothesis on synthetic datasets where each of the two classes is generated by a Gaussian distributions, i.e., $D_1,D_2$, in $\mathbb{R}^{20}$.
Note that for Gaussian, KL-divergence can be analytically computed as the closed form exists.
In this case, Figure~\ref{fig:kld_gauss} shows that \padv \emph{significantly outperforms} all the baselines by a larger margin when the class closeness is small. 
That is, for datasets that are challenging to classify, \padv outperforms the competition.

Furthermore, we revisit the real datasets from Table~\ref{tab:main} and re-interpret the results under the lens of their class closeness. 
Since the true data distributions of these datasets are unknown, we approximate the KL-divergence via data samples using the estimator in Equation~(5) from \citet{kld}. 
Indeed, Figure~\ref{fig:kld_others} shows that the datasets on which \padv clearly outperforms \yang and \wang, are the ones with small class closeness (e.g., Diabetes, Gaussian).

\subsection{Runtime Comparisons} \label{ssec:runtime}

\begin{figure}[t]
    \centering
    \hfill
    \begin{subfigure}[t!]{0.31\textwidth}
        \centering
        \includegraphics[width=\textwidth]{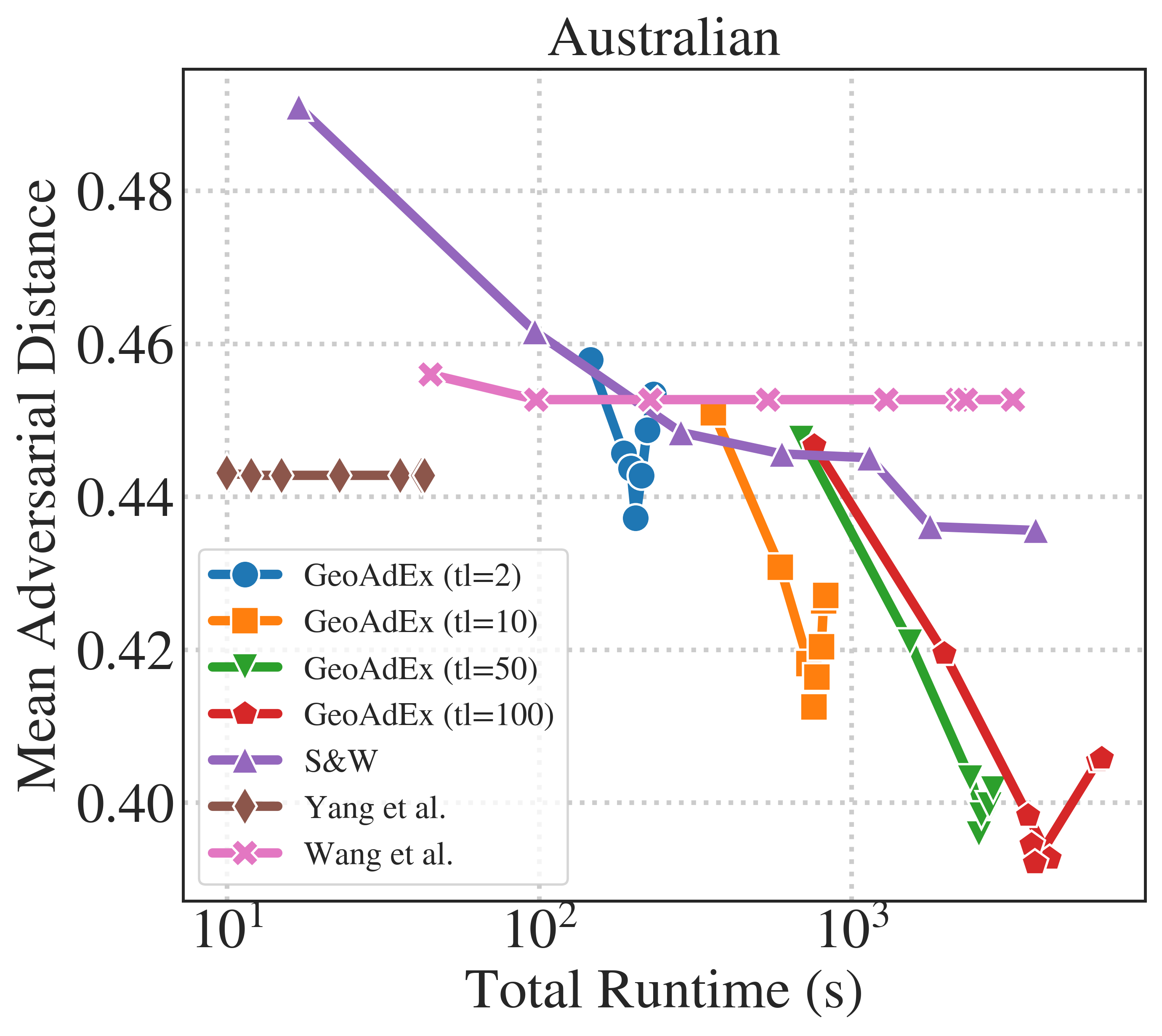}
        \caption{Australian}
        \label{fig:runtime_australian}
    \end{subfigure}
    \hfill
    \begin{subfigure}[t!]{0.31\textwidth}
        \centering
        \includegraphics[width=\textwidth]{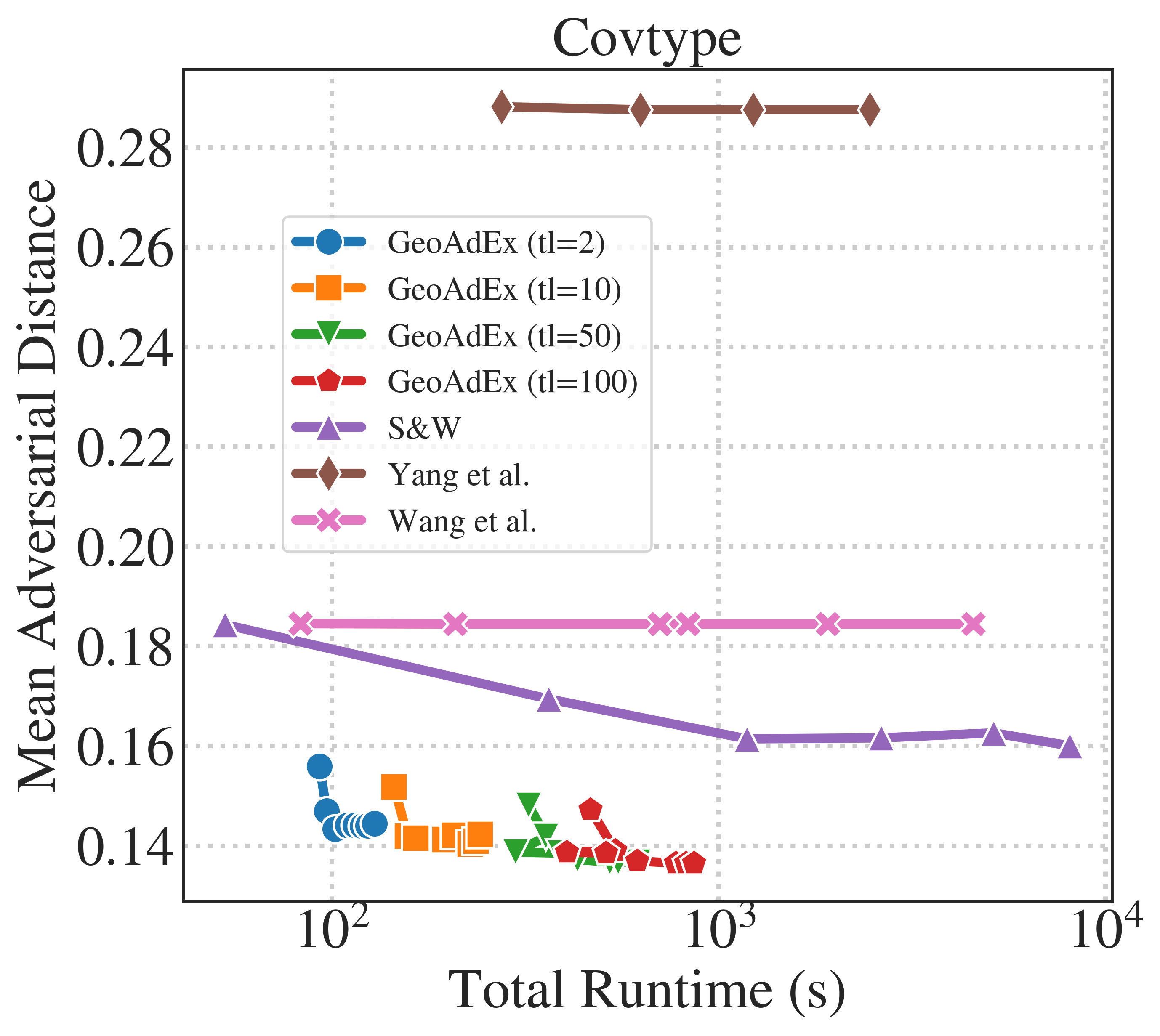}
        \caption{Covtype}
        \label{fig:runtime_covtype}
    \end{subfigure}
    \hfill
    \begin{subfigure}[t!]{0.31\textwidth}
        \centering
        \includegraphics[width=\textwidth]{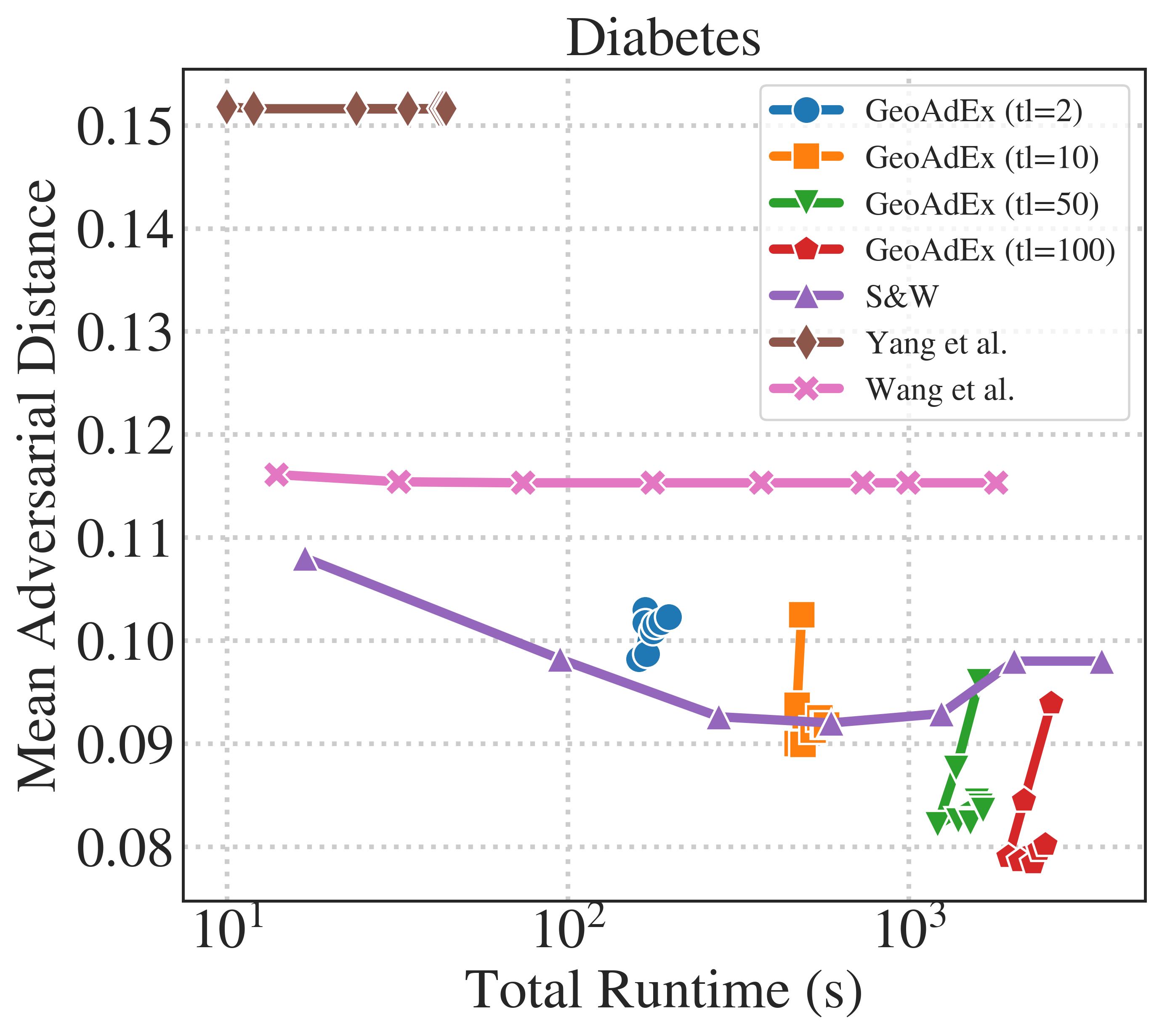}
        \caption{Diabetes}
        \label{fig:runtime_diabetes}
    \end{subfigure}
    \hfill~
    \caption{Mean adversarial distance vs. total runtime for \padv and all the baselines with different choices of hyperparameters on (a) Australian, (b) Covtype, and (c) Diabetes. Each point represents a single run with a unique set of hyperparameters. For a given runtime, an attack with the lowest mean adversarial distance is the best. \padv utilizes the extra runtime better and outperforms the baselines.}
    \label{fig:runtime}
    \vspace{-10pt}
\end{figure}

Runtimes of the experiments in Table~\ref{tab:main} are reported in Table~\ref{tab:runtime}.
Through the choice of hyperparameters, we attempt to control the runtime of each algorithm to roughly be within the same order of magnitude, but this has proven difficult given that we do not wish to fine-tune the hyperparameters specifically per dataset.
To better compare the algorithms under a similar set of runtimes, we repeat the experiments with different sets of hyperparameters chosen at certain intervals and plot the mean adversarial distance vs. the total runtime for each experiment in Figure~\ref{fig:runtime}. 

For each algorithm, we started with its default hyperparameters and adjusted them (either linearly or exponentially) in the direction that should find smaller adversarial perturbations with a longer runtime.
Each point on the curves represents an experiment with one set of hyperparameters.
One way to read the plot is to fix a particular runtime and compare the mean adversarial distance from each line.
This experiment is expensive so we only run it for the first three datasets from Table~\ref{tab:main}.

For Australian and Diabetes, in a regime with very short runtime (${\sim}100$s total or ${\sim}1$s per sample), there is at least one baseline that is both faster and finds smaller adversarial perturbation than \padv. 
However, with longer runtimes, our \padv outperforms all the baselines by a large margin ($10\%$, $15\%$ and $20\%$ improvement over the second best for Australian, Covtype, and Diabetes, respectively). 
In most settings, the baselines do not benefit much, if at all, from the increased runtime. 
Conversely, \padv always finds smaller adversarial distances given a longer time limit.
More details and discussions of this experiment are in Appendix~\ref{app:runtime}.

\textbf{Effect of the initialization attack.}
The initialization of $\epsilon$ with \sit attack, as described in the performance optimization (II), also improves the total runtime of \padv. 
Depending on the dataset, the runtime reduction range from $2\%$ to $75\%$.
In the interest of space, ablation study and analysis of the initialization attack is included in Appendix~\ref{app:ablation}.

\section{Discussion \& Open Problems} \label{sec:discussion}

\padv has been shown to be successful in discovering adversarial examples for $k$-NN classifiers with $k \geq 1$. 
Compared to the baselines, it finds adversarial examples with a considerably smaller perturbation norm on most of the commonly tested datasets. 

The main limitation of \padv is its runtime, particularly in the short-runtime regime.
\sit is always the fastest, generally followed by \wang, \yang, and then \padv. 
However, in the long-runtime regime, our \padv outperforms the baselines by a large margin given the same runtime.
\padv utilizes a longer runtime by expanding the search radius while the baselines search more cells without a particular order or prioritization. 
As a result, the adversarial distance found by \sit, \yang, and \wang typically does not improve much further with more computation.

We note that compared to the geometric approach by \citet{jordan19geocert}, our experiments were run on a significantly larger scale. 
Specifically, \citet{jordan19geocert} was tested on neural networks with only up to 70 ReLUs, which are much smaller than typical networks used in practice. This is equivalent to Voronoi cells with only 70 neighbors/facets in our setting. \padv was tested on datasets with over ten thousand generator points and $k$ up to 7, which results in a substantially larger number of polytopes to search through. 

\padv has shown convincing results with our new geometric take on the problem, but there is room for improvement in terms of efficiency. 
Additional speedups can be introduced via parallelization, GPU utilization, and a faster optimization technique. 
A more sophisticated heuristic for determining the order-1 neighbors can also be used. 
Furthermore, geometric properties of a high-order Voronoi diagram may be better exploited to save unnecessary computation on non-adversarial cells. 
For instance, multiple neighboring non-adversarial cells may be combined and approximated as a single large cell if we can guarantee that there is no adversarial cell inside this new polytope. 
This would remove a large number of distance computations surrounding the given test point, which is the main bottleneck of \padv. 
Additionally, \padv can also be extended to other space-partitioning classifiers such as decision trees and random forests.
\section{Conclusion}

We propose \padv, an algorithm based on geometric insights for finding adversarial examples on $k$-NN classifiers. 
We leverage the structural properties of higher-order Voronoi diagrams to propose efficient approximations and speedup the final algorithm. 
While \padv typically requires a higher computational cost, it  significantly outperforms the baselines in most of the datasets. 
Finally, for the case where there is no clear separation between the classes (a typical characteristic of real data) \padv significantly   outperforms the competition.  

\subsubsection*{Acknowledgement}
The first author of this paper was supported by the Hewlett Foundation through the Center for Long-Term Cybersecurity (CLTC) and by generous gifts from Open Philanthropy and Google Cloud Research Credits program with the award GCP19980904.

The second and third authors were supported by the Center for Long-Term Cybersecurity (CLTC), the Berkeley Deep Drive project, NSF grant TWC-1518899, and Open Philanthropy.

  {\small
    \bibliographystyle{abbrvnat}
    \bibliography{bibliography/ml,bibliography/advex,bibliography/advml,bibliography/knn}
  }

\clearpage

\appendix
\onecolumn
\section*{\Large Appendix}

Appendix~\ref{ap:exp_detail} contains details on the experiments conducted throughout this paper. 
In Appendix~\ref{ap:add_results}, we include additional results from the experiments on \padv, e.g., attacks on $k$-NN with $k=1$, runtime comparisons, an ablation study, and different hyperparameter choices.
In Appendix~\ref{ap:closeness}, more details on the class closeness metric are provided.
In Appendix~\ref{ap:proof}, we provide the proofs of the theory and lemmas stated in the paper. 
Appendix~\ref{ap:opt} explains all the performance optimization used in \padv, and lastly, in Appendix~\ref{ap:dual}, we describe the optimization algorithm, greedy coordinate ascent, used for the distance computation.

\section{Details of the Experiments} \label{ap:exp_detail}

\paragraph{Datasets.} Details regarding the datasets used in the experiments are included in Table~\ref{tab:datasets}. It also includes the accuracy of $k$-NN classifiers at $k=1,3,5,7$. Australian, Covtype, Diabetes, Fourclass, and fmnist06 are taken directly from \yang's implementation. The dataset fmnist06 is a two-class subset of Fashion-MNIST with a dimension reduction to 25 via PCA. The Letters dataset, together with the others, is taken from LIBSVM \citep{libsvm}\footnote{\url{https://www.csie.ntu.edu.tw/~cjlin/libsvmtools/datasets/multiclass.html}}. Gaussian is a dataset we create by sampling from two isotropic Gaussian distributions of 20 dimension and variance of 1. The distance between the means of the two distributions is 1 by default and is varied only in Section~\ref{ssec:structure} to get different values of class closeness.

\paragraph{Environment and implementation.} All of the attacks are run on an Ubuntu (16.04) cluster with 128 AMD EPYC 7551 CPU cores (2.5GHz each) and 252 GB of memory. No GPU is used in any of the experiments. Using GPU could further speed up the attacks, but the official implementation of the baselines is not compatible with GPUs so we stick to CPUs to present a fair comparison. \yang uses explicit parallelization and Cython C-extensions, whereas the rest of the attacks use pure Python code without explicit parallelization. The code for the baselines are taken directly from their respective public repository.\footnote{\sit: \url{https://github.com/chawins/knn-defense}, \yang: \url{https://github.com/yangarbiter/adversarial-nonparametrics}, \wang: \url{https://github.com/wangwllu/knn_robustness}} \yang uses Gurobi as the solver.

\paragraph{Hyperparameters.} We evaluate all the baselines using their publicly available code and default hyperparameters. 
For fairness, we also attempt to tune the hyperparameters for each baseline to keep the total runtime comparable across attacks (see Appendix~\ref{ssec:runtime}).
In general, the baselines are fairly insensitive to changes in their hyperparameters.
For example, increasing the number of Voronoi cells searched by \yang and \wang almost never reduces the mean perturbation norm beyond one obtained with the default value.
For \padv, we choose to compute distance to cell and set $m$ to $20$ and applied a time limit of $100$ seconds per test point. 

\begin{table*}[!ht]
\small
\centering
\vspace{-5pt}
\caption{Details of the datasets used in the experiments.}
\label{tab:datasets}
\begin{tabular}{@{}lrrrrrrr@{}}
\toprule
Datasets   & \# points & \# features & \# classes & $k=1$ acc & $k=3$ acc & $k=5$ acc & $k=7$ acc \\ \midrule
Australian & 490       & 14          & 2          & 0.805          & 0.805     & 0.830     & 0.845  \\
Covtype    & 2000      & 54          & 7          & 0.755          & 0.715     & 0.730     & 0.705   \\
Diabetes   & 568       & 8           & 2          &  0.695         & 0.755     & 0.695     & 0.685     \\
Fourclass  & 662       & 2           & 2          &  1.000         & 0.995     & 1.000     & 1.000     \\
Gaussian   & 10000     & 20          & 2          &  0.550         & 0.660     & 0.640     & 0.635     \\
Letters    & 15000     & 16          & 26         &   0.925        &  0.940    & 0.940     & 0.930   \\
fmnist06   & 12000     & 25          & 2          &   0.800        & 0.795     &  0.810         &  0.810         \\ \bottomrule
\end{tabular}
\vspace{-10pt}
\end{table*}

\section{Additional Results} \label{ap:add_results}

\subsection{Exact Attacks for $k=1$} \label{app:k1}

\begin{table*}[t]
\small
\centering
\caption{Runtime for the exact version of the attacks on all of the datasets with $k=1$. \sit is not included because it does not offer an exact solution and provide no guarantee on the adversarial distance.}
\label{tab:k1}
\begin{tabular}{@{}llllllll@{}}
\toprule
Attacks & Australian & Covtype      & Diabetes & Fourclass & Gaussian & Letters & fmnist06 \\ \midrule
\yang & 72 & 7186 & 66 & 59 & 9109 & 34997 & 18461 \\
\wang & 2 & 10 & 2 & 25 & 159 & 78 & 333 \\
\textbf{\padv} & 17 & 38 & 15 & 39 & 476 & 7450 & 19307 \\ \bottomrule
\end{tabular}
\end{table*}

For completeness, we also compare the exact version of the attacks on $1$-NN where the results are presented in Table~\ref{tab:k1}. Note that \sit is excluded since it does not have an exact version. 
\wang is generally the fastest, and \padv is faster than \yang, which does not seem to scale well with the number of generators and dimension. This is a direct effect of the solvers of the quadratic programs. Greedy coordinate ascent, used by \wang and our attack, is much more efficient than a general-purpose commercial solver.

\subsection{Runtime Comparisons and Attack Hyperparameters} \label{app:runtime}

\begin{table*}[t]
\small
\centering
\caption{Runtime (in seconds) for each of the attacks on all of the seven datasets. The mean adversarial distance corresponding to these runtimes are shown in Table~\ref{tab:main}. The numbers in the gray rows are 95\%-confidence interval from 10 runs with random splits between training and testing samples.}
\label{tab:runtime}
\begin{tabular}{@{}llrrrrrrr@{}}
\toprule
$k$                & Attacks          & Australian & Covtype      & Diabetes & Fourclass & Gaussian & Letters & fmnist06 \\ \midrule
& S\&W [2020] & $654$ & $1225$ & $465$ & $336$ & $811$ & $3372$ & $972$ \\
\rowcolor{Gray}
\cellcolor{white} & \cellcolor{white} & $\pm 1$ & $\pm 250$ & $\pm 121$ & $\pm 5$ & $\pm 14$ & $\pm 13$ & $\pm 179$ \\
& \yang & $8$ & $925$ & $9$ & $7$ & $599$ & $1555$ & $2151$ \\
\rowcolor{Gray}
\cellcolor{white} & \cellcolor{white} & $\pm 1$ & $\pm 97$ & $\pm 1$ & $\pm 1$ & $\pm 6$ & $\pm 38$ & $\pm 62$ \\
& \wang & $361$ & $363$ & $128$ & $129$ & $226$ & $259$ & $272$  \\
\rowcolor{Gray}
\cellcolor{white} & \cellcolor{white} & $\pm 21$ & $\pm 20$ & $\pm 9$ & $\pm 5$ & $\pm 19$ & $\pm 19$ & $\pm 21$ \\
& \textbf{\padv} & $3351$ & $727$ & $1987$ & $1424$ & $2525$ & $4030$ & $6427$ \\
\rowcolor{Gray}
\multirow{-8}{*}{\cellcolor{white}3} & \cellcolor{white} & $\pm 447$ & $\pm 100$ & $\pm 152$ & $\pm 106$ & $\pm 146$ & $\pm 354$ & $\pm 338$ \\
\cmidrule{2-9}
& S\&W [2020] & $654$ & $1225$ & $465$ & $336$ & $811$ & $3372$ & $972$ \\
\rowcolor{Gray}
\cellcolor{white} & \cellcolor{white} & $\pm 1$ & $\pm 250$ & $\pm 121$ & $\pm 5$ & $\pm 14$ & $\pm 13$ & $\pm 179$ \\
& \yang & $8$ & $925$ & $9$ & $7$ & $599$ & $1555$ & $2151$ \\
\rowcolor{Gray}
\cellcolor{white} & \cellcolor{white} & $\pm 1$ & $\pm 97$ & $\pm 1$ & $\pm 1$ & $\pm 6$ & $\pm 38$ & $\pm 62$ \\
& \wang & $361$ & $363$ & $128$ & $129$ & $226$ & $259$ & $272$  \\
\rowcolor{Gray}
\cellcolor{white} & \cellcolor{white} & $\pm 21$ & $\pm 20$ & $\pm 9$ & $\pm 5$ & $\pm 19$ & $\pm 19$ & $\pm 21$ \\
& \textbf{\padv} & $3351$ & $727$ & $1987$ & $1424$ & $2525$ & $4030$ & $6427$ \\
\rowcolor{Gray}
\multirow{-8}{*}{\cellcolor{white}5} & \cellcolor{white} & $\pm 447$ & $\pm 100$ & $\pm 152$ & $\pm 106$ & $\pm 146$ & $\pm 354$ & $\pm 338$ \\
\cmidrule{2-9}
& S\&W [2020] & $654$ & $1225$ & $465$ & $336$ & $811$ & $3372$ & $972$ \\
\rowcolor{Gray}
\cellcolor{white} & \cellcolor{white} & $\pm 1$ & $\pm 250$ & $\pm 121$ & $\pm 5$ & $\pm 14$ & $\pm 13$ & $\pm 179$ \\
& \yang & $8$ & $925$ & $9$ & $7$ & $599$ & $1555$ & $2151$ \\
\rowcolor{Gray}
\cellcolor{white} & \cellcolor{white} & $\pm 1$ & $\pm 97$ & $\pm 1$ & $\pm 1$ & $\pm 6$ & $\pm 38$ & $\pm 62$ \\
& \wang & $361$ & $363$ & $128$ & $129$ & $226$ & $259$ & $272$  \\
\rowcolor{Gray}
\cellcolor{white} & \cellcolor{white} & $\pm 21$ & $\pm 20$ & $\pm 9$ & $\pm 5$ & $\pm 19$ & $\pm 19$ & $\pm 21$ \\
& \textbf{\padv} & $3351$ & $727$ & $1987$ & $1424$ & $2525$ & $4030$ & $6427$ \\
\rowcolor{Gray}
\multirow{-8}{*}{\cellcolor{white}7} & \cellcolor{white} & $\pm 447$ & $\pm 100$ & $\pm 152$ & $\pm 106$ & $\pm 146$ & $\pm 354$ & $\pm 338$ \\
\bottomrule
\end{tabular}
\end{table*}

Table~\ref{tab:runtime} includes the runtime of all the attacks for $k = 3,5,7$. 
As we have mentioned, \padv with $m = 20$ and the time limit of 100 seconds takes longer to run compared to the other attacks for most cases. 
\sit is the fastest, followed by \wang and \yang. 
For the other cases, \yang has the longest runtime. 
The reported runtime of \padv also includes the time used to initialize the adversarial distance $\epsilon$ found by \sit.

As mentioned in Section~\ref{ssec:runtime}, we conduct more thorough sets of experiments on the first three datasets (Australian, Covtype, and Diabetes) to fairly compare the attacks under a similar runtime.
To this end, we plot the runtime vs. mean adversarial distance curves for \padv and all the baselines by varying their hyperparameters in Figure~\ref{fig:runtime}.

\subsubsection{Runtime Experiment Setup}
For \padv, we vary $m$ with four different values of time limit per sample which result in four different curves.
For \yang, we progressively doubled the number of regions searched which is the only adjustable hyperparameter. 
For \wang, we progressively doubled the number of trials (both min and max) and the number of neighbors to consider (until it exceeds the number of all generators).
There are four hyperparameters for \sit: \texttt{binary\_search\_steps}, \texttt{max\_iterations}, \texttt{thres\_steps}, and \texttt{check\_adv\_steps}. 
We progressively increased the first two and decreased the last two linearly. For \padv, we tested a more detailed breakdown by varying both  (from $5$ to $120$) and time limit. 

Note that the runtime can be slightly different from what reported in Table~\ref{tab:runtime} since we use a different machine. 
To make this figure, we run all the experiments (all attacks, datasets, and choices of hyperparameters) on a server with 40 cores of Intel(R) Xeon(R) Gold 6230 CPU @ 2.10GHz.

\subsubsection{Discussion on Figure~\ref{fig:runtime}}

The major trends have already been covered in Section~\ref{ssec:runtime}.
Here, we discuss other observations and minor trends.

\textbf{Increasing $\bm{m}$ in \padv can either increase or decrease the mean adversarial distance.}
Each curve for \padv is generated by increasing $m$ but fixing the time limit. 
Increasing $m$ reduces the chance of missing the adversarial facets (hence the downward trend in the adversarial distance), but it also increases the computation time for each cell which means that there are fewer cells it can search given a fixed time limit (hence the upward trend).
This implies that there is an optimal value of $m$ for a given time limit.

\textbf{Increasing $\bm{m}$ in \padv can either increase or decrease the total runtime.}
This outcome is seemingly perplexing than the previous one.
We explain it for different values of $m$, namely the small-$m$ and the large-$m$ regions. 

\emph{Small-$m$ region}. 
When a smaller $m$ is used with \padv, fewer first-order neighbors are considered, and thus, the search has a higher chance of missing facets and nearby adversarial cells completely. 
As a result, it has to expand the search radius which, in turn, would discover adversarial examples that are further away and use a longer runtime. 
Conversely, if we increase $m$, we may find these previously missed cells and terminate earlier, resulting in both lower adversarial distance and runtime. 
We call the first scenario from the above exposition, the ``small-$m$ region.''

\emph{Large-$m$ region}. 
On the other hand, when $m$ is sufficiently large and no adversarial cell is missed, increasing $m$ could have a reversed effect. 
In particular, when $m$ increases, more first-order neighbors have to be considered, and hence more \emph{nearby} cells will have to be searched. 
For each of the test samples, this could lead to (i) an increased runtime and/or (ii) the previously found adversarial cells that are further away may now be missed instead. 
When (ii) happens, \padv will timeout and just return the initialized upper bound. 
Therefore, in the ``large-$m$ region,'' both the adversarial distance and the total runtime may increase with $m$. 

We verify this hypothesis by inspecting the number of samples that are timed out by \padv. 
If our hypothesis holds, when we test different values of $m$, we expect to see a decreasing trend on the number of timeouts in the small-$m$ region and an increasing trend in the large-$m$ region. 
Specifically, when varying $m \in \{5,10,20,40,60,80,100,120\}$ on the Diabetes dataset, we observe the following number of timeouts over 100 samples: 41, 21, 14, 16, 10, 15, 16, and 17. 
The first three experiments ($m=5,10,20$) which correspond to the small-$m$ region show a decreasing number of timeouts from 41 to 14. 
The last three experiments ($m=80,100,120$) correspond to the large-$m$ region where both runtime and distance increase with $m$. 
The same phenomenon also happens on Covtype given the same hyperparameters and the range of values of $m$. 
In this case, the numbers of timeouts are 24, 9, 3, 3, 3, 4, 4, and 5, respectively.

Note that whether a value of $m$ is considered ``small'' or ``large'' varies by datasets and the time limit. Increasing the time limit reduces the number of timeouts and hence delays the large-$m$ region (i.e., occurs at a larger $m$). 
Additionally, this observation leads to two practical suggestions: (1) regardless of $m$, increasing the time limit is always beneficial in terms of the adversarial distance, but (2) for a fixed time limit, there is an optimal value of $m$.

\textbf{Curves for \yang are shorter than the others.}
For Australian and Diabetes, the lines associated with \yang are shorter than the rest because we cannot increase the total runtime by adjusting the hyperparameter any further. 
This is a fundamental flaw of the heuristic used by \yang which only searches the cells that contain any generator from a wrong class. 
So the total number of cells searched is upper bound by the number of generators from a wrong class which is very limited compared to the total number of cells.

\subsection{Ablation Study} \label{app:ablation}

\begin{figure}[t]
    \centering
    \begin{minipage}{0.46\textwidth}
        \centering
        \includegraphics[width=\textwidth]{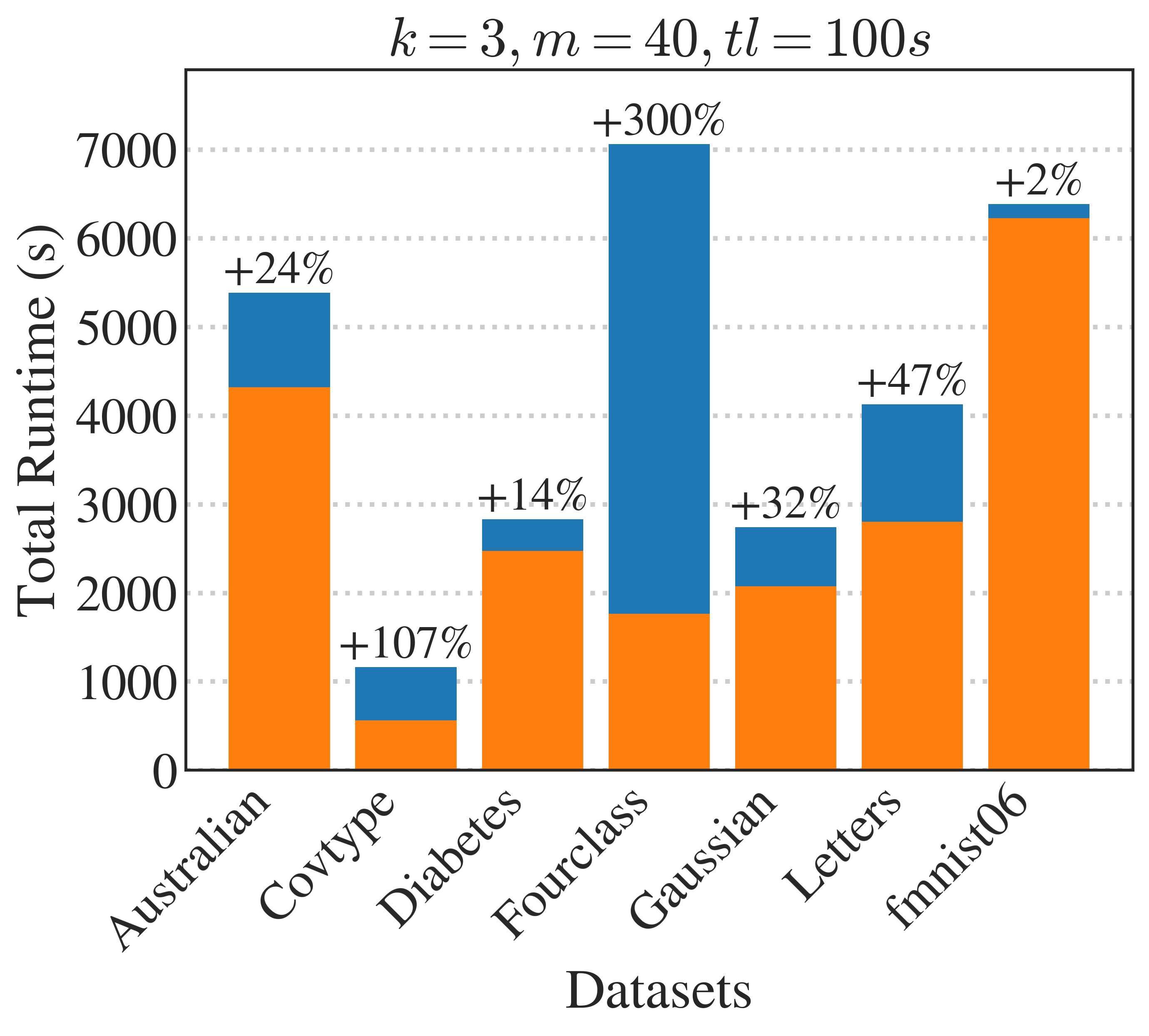}
        \caption{Improvement in the total runtime of \padv with (orange) and without (blue) \sit initialization.}
        \label{fig:sw_init}
    \end{minipage}
    \hfill
    \begin{minipage}{0.46\textwidth}
        \centering
        \includegraphics[width=\textwidth]{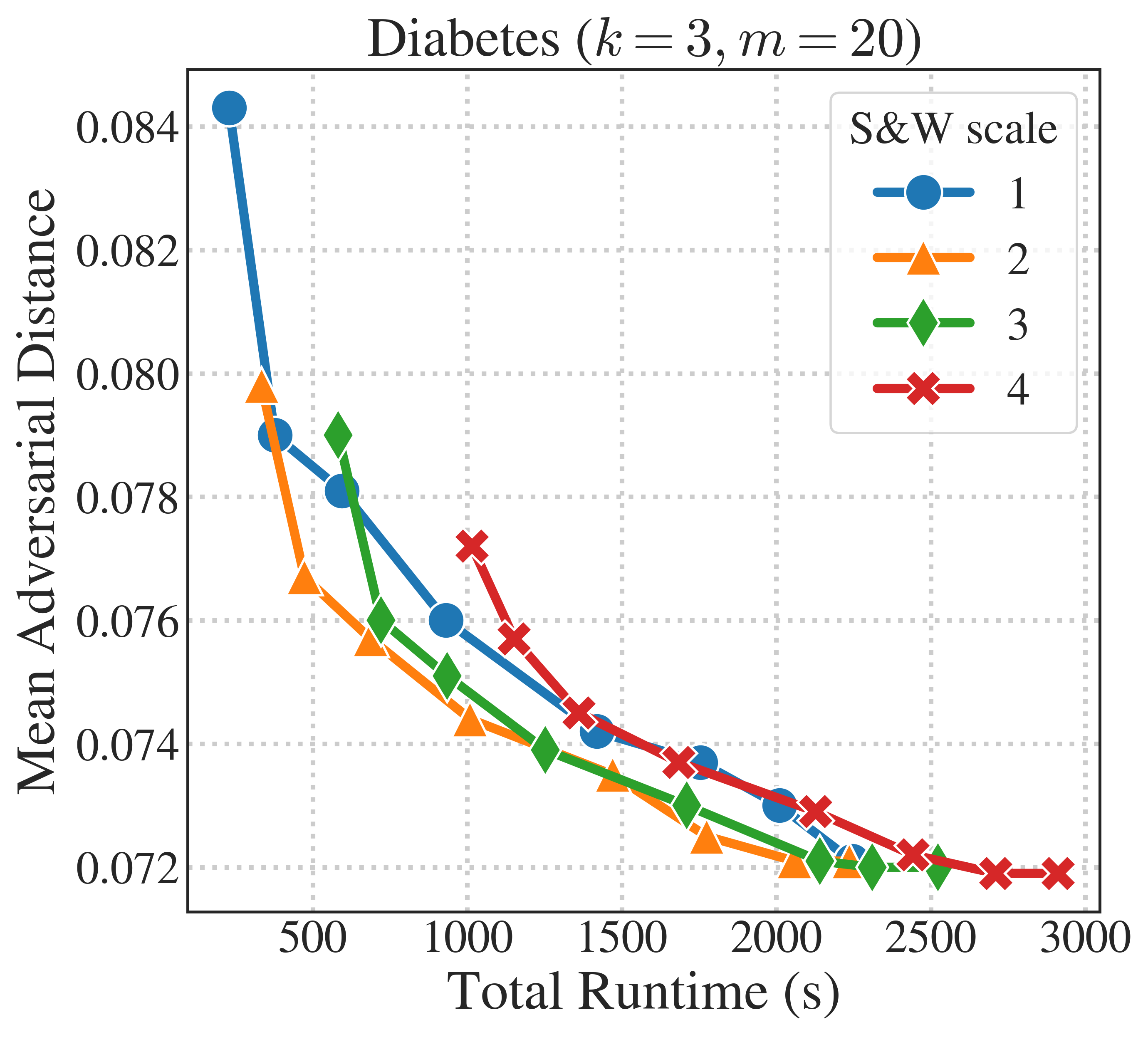}
        \caption{Mean adversarial distance vs. total runtimes on \padv using S\&W initialization with four different hyperparameter scaling (corresponding to the four curves). Each point on the curve represents a unique choice of the time limit per sample in \padv (from 5 to 200 seconds).}
        \label{fig:sw_scale}
    \end{minipage}
    \vspace{-10pt}
\end{figure}

\subsubsection{Importance of S\&W Initialization}

We want to compare \padv with and without the S\&W initialization.
Figure~\ref{fig:sw_init} compares the total runtime of \padv on all of the datasets with $k=3$ and shows that the S\&W initialization speeds up the attack in all cases.
While effects of the initialization are generally minor, we found two cases, namely Covtype and Fourclass, where the initialization leads to a large improvement in the runtime. 
Without the initialization, the runtime increases by $107\%$ and $300\%$ in these two datasets respectively. 
We give our explanation for these two datasets below.

For Covtype, \padv's runtimes are 561s with the initialization step and 1163s without. 
Without an initial upper bound on $\epsilon$, there is a small chance that \padv misses a nearby adversarial cell and keeps running until the time limit is met.
While only several test samples are affected, it ends up raising the total runtime by a relatively large margin ($100\%$ from ~500s) because the time limit is set to 100s per sample. 
The initialization has the same kind of effect on Fourclass as on Covtype, but it affects a much larger number of samples. 
The number of timeouts goes from 4 to 43 as the initialization is removed, explaining the significant increase in the total runtime. 
One hypothesis is that this phenomenon takes place because S\&W attack is particularly effective on Fourclass (note its competitive performance compared to the other attacks), and thus removing it as an initialization results in an equally large degradation on the performance of \padv.

\subsubsection{Effects of S\&W Initialization's Hyperparameters}

To test the effect of the initialization’s hyperparameters on \padv, we ran a simple experiment on the Diabetes dataset by varying the runtime of S\&W initialization and the time limit of \padv. 
The trade-off curves between total runtime and the adversarial distance are shown in Figure~\ref{fig:sw_scale}. 
Each curve uses a different scaling factor of the S\&W attack as described in our previous experiment to vary its runtime. 
Each dot in the curve is generated by varying the time limit of \padv from 5 to 200 seconds. For a wide range of total runtimes, S\&W initialization with a scaling factor between $2$ and $3$ (or anywhere between $4\%$ and $35\%$ of the total runtime) seems like an optimal spot. 
This suggests that performance of \padv is fairly insensitive to the hyperparameters of the S\&W initialization. 
Based on this observation alone, one may aim to run the initialization for ${\sim}10{-}20\%$ of the total runtime as the simplest baseline and avoid any additional overhead.

We simply choose S\&W attack because it finds an adversarial example quickly (even though the upper bound is relatively loose) with little impact on the total runtime. 
The choice is somewhat arbitrary and can be replaced with the attacks by \yang or \wang.
Deciding how long to run S\&W initialization is a good question from a practical standpoint. 
It also depends on many factors including the dataset, the desired time limit, and hyperparameters of \padv. 
So it should be considered on a case-by-case basis, perhaps by a hyperparameter optimization on a small subset of the data. 
A cheaper strategy that works across all datasets is to terminate the initialization attack (per-sample) when the adversarial distance starts to plateau (e.g, less than $A\%$ improvement in the last $B$ iterations).

\section{Class Closeness} \label{ap:closeness}

Intuitively, the \emph{class closeness} measures distance between one distribution to another that has a different class and is closest to it. We only consider the closest class because when generating an adversarial example, one only has to perturb the test point towards the nearest distribution with a different class and the other classes are almost irrelevant. More formally, we can write the class closeness as
\begin{align*}
    \text{class closeness} ~\coloneqq~ \frac{1}{c} \sum_{i=1}^c ~\min_{j \in \{1,\dots,c\}} \textsf{KLD}(D_i || D_j)
\end{align*}
where $c$ is the number of classes, and $D_i$ is the distribution conditioned on class $i$.

We first experiment with the Gaussian dataset because its KL-divergence has a analytical form. Specifically, the KL-divergence between two multivariate Gaussian distributions in $\mathbb{R}^d$, $D_1 = \mathcal{N}(\mu_1, \Sigma_1)$ and $D_2 = \mathcal{N}(\mu_2, \Sigma_2)$, is given by
\begin{align*}
    \textsf{KLD}(D_1 || D_2) = \frac{1}{2} \bigg[ &\log \frac{\abs{\Sigma_2}}{\abs{\Sigma_1}} - d + \text{tr}(\Sigma_2^{-1} \Sigma_1) + (\mu_2 - \mu_1)^\top \Sigma_2^{-1} (\mu_2 - \mu_1) \bigg]
\end{align*}

In particular, we use isotropic Gaussian distributions so the means and the covariance matrices can be simplified even further. 
\begin{align*}
    \mu_1 &= \begin{bmatrix} \alpha\\ 0\\ \vdots\\ 0 \end{bmatrix},
    \mu_2 = \begin{bmatrix} -\alpha\\ 0\\ \vdots\\ 0 \end{bmatrix},
    \Sigma_1 = \Sigma_2 = I_d
\end{align*}
Note that we pick $d = 20$, and without loss of generality, we can simply assign different values of $\alpha$ and $-\alpha$ to the first coordinate of $\mu_1$ and $\mu_2$ to vary the distance between the two means. We pick $\alpha$ among $\{0.3, 0.5, 0.7, 0.9, 1.1, 1.3, 1.5\}$. This specific case yields a very simple form of KL-divergence: \begin{align*}
\textsf{KLD}(D_1 || D_2) &= 2\alpha^2
\end{align*}

For the second part, since the distributions of the other datasets are unknown, we use a non-parametric method from \citet{kld} to approximate the KL-divergence. This method only requires samples from the distributions and is coincidentally based on $k$-NN. We pick $k = 5$ for this approximation method which has nothing to do with the value of $k$ in $k$-NN classifiers we experiment with.

\section{Proofs} \label{ap:proof}

\subsection{Theorem~\ref{thm:firstorder}}

Now we restate Theorem~\ref{thm:firstorder} and then the proof.
\firstorder*

\begin{proof}
Let $V(x'|G)$ denote the order-1 Voronoi cell for $x'$ on the set of generators $G$. 
From property OK1 in Section 3.2.1 of~\cite{10.5555/135734} we know that the order-$k$ Voronoi cell $V(S\cup\{x_i\})$ can be expressed as:
\begin{align*}
V(S\cup\{x_i\}) =& \left(\bigcap\limits_{l=1}^{k-1}V(x_l|(X\setminus (S \cup \{x_i\}))\cup\{x_l\})\right) \cap V\left(x_i|X\setminus S \right)
\end{align*}

From the fact that $V(S\cup\{x_i\})$ is a order-$k$ Voronoi cell, we know that $V(S\cup\{x_i\})$ is nonempty. 
Let us assume for the sake of contradiction that $V(\{x_i\})$ is not neighboring with any of the $V(\{x_1\}),\ldots,V(\{x_{k-1}\})$. 
Then we know that: 

\[V(\{x_i\}) = V\left(x_i|X\setminus S \right)\]

This is because the removal of $S$ from the set of generators of the Voronoi diagram did not affect $V(\{x_i\})$ since $V(\{x_i\})$ is not neighboring with any of $V(\{x_1\}),\ldots,V(\{x_{k-1}\})$. 
Additionally, the removal of $x_i$ from the set of generators in the term $V\left(x_l|x_l\cup \left(X\setminus S \cup \{x_i\}\right)\right)$ is not affecting the corresponding cell, again, because  $V(\{x_i\})$ is not neighboring with  $V(\{x_1\}),\ldots,V(\{x_{k-1}\})$. 
This implies that:
\[V(x_l|(X\setminus (S	 \cup \{x_i\}))\cup\{x_l\})=V(x_l|(X\setminus S	 )\cup\{x_l\})\]

Using the above observations we can rewrite the first relation as:
\begin{align*}
V(S\cup\{x_i\})=\left(\bigcap\limits_{l=1}^{k-1} V(x_l|(X\setminus S	 )\cup\{x_l\}) \right)\cap V\left(\{x_i\} \right)
\end{align*}

For the last part of the proof we will show that the above intersection is empty which contradicts the fact that $V(S\cup\{x_i\})$ is a nonempty Voronoi cell. 
Notice that the changes in the set of generators that take place in the term $V(x_l|(X\setminus S)\cup\{x_l\})$ for $l=[1,k-1]$ do not affect the Voronoi cell of $x_i$. 
This means that even after the changes in the set of generators, the cell of $x_i$ is a superset of $V(\{x_i\})$, or to put it differently, the polytope $V(x_l|(X\setminus S	 )\cup\{x_l\})$ never enters the area of  $V(\{x_i\})$.
As a result, none of these terms intersects with $V(\{x_i\})$.
But this contradict the fact that $V(S\cup\{x_i\})$ is nonempty.

\end{proof}

\subsection{Lemma~\ref{thm:correct}: Correctness of \padv}

\correct*

Lemma~\ref{thm:correct} can be obtained directly from a similar theorem in \citet{jordan19geocert}. We first restate this theorem and the definition of \emph{polyhedral complex}. Then, we provide a short proof.

\begin{restatable}[Correctness of GeoCert]{thm}{geocert}
\label{thm:geocert}
(Restate from Theorem C.2 in \citet{jordan19geocert}) For a fixed polyhedral complex $\mathscr{P}$, a fixed input point $x_0$ and a potential function $\phi$ that is ray-monotonic, GeoCert returns a boundary facet with minimal potential $\Phi$.
\end{restatable}

\begin{restatable}[Polyhedral Complex]{definition}{polyhedral}
\label{thm:polyhedral}
(Restate from Definition 2 in \citet{jordan19geocert}) A nonconvex polytope, described as the union of elements of the set $\mathscr{P} = {\mathcal{P}_1, ...,\mathcal{P}_k}$ forms a polyhedral complex if, for every $\mathcal{P}_i, \mathcal{P}_j \in \mathscr{P}$ with nonempty intersection, $\mathcal{P}_i \cap \mathcal{P}_j$ is a face of both $\mathcal{P}_i$ and $\mathcal{P}_j$.
\end{restatable}

\begin{proof}
In order to apply Theorem~\ref{thm:geocert} to \padv, we need to show two things: (i) the test point lies in a polyhedral complex, and (ii) Euclidean distance is a ray-monotonic potential function. 
First, it is trivial to see that the set of non-adversarial Voronoi cells connected to and including the cell the test input $x$ falls into forms a polyhedral complex. Since Voronoi cells are polytopes and any pair of them intersect at most at the shared facet, any set of Voronoi cells forms a polyhedral complex. This (informally) proves part (i).

For part (ii), we refer the readers to Corollary C.3 of \citet{jordan19geocert} which shows that Euclidean distance is a ray-monotonic potential function. With these two conditions in mind, \padv behaves in the same way as GeoCert algorithmically in their respective settings, and so Theorem~\ref{thm:geocert} directly applies to \padv as well.
\end{proof}

\subsection{Lemma~\ref{thm:lowerbound}: Lower Bound of the Optimal Adversarial Distance}

\lowerbound*

\begin{restatable}{thm}{geocert2}
\label{thm:geocert2}
(Restate from Lemma C.1 in \citet{jordan19geocert}) For any polyhedral complex $\mathscr{P}$ point $x_0$, and ray-monotonic potential $\phi$, let $\mathcal{F}_i$ be the
facet popped at the $i$-th iteration of GeoCert. Then for all $i < j, \Phi(\mathcal{F}_i) < \Phi(\mathcal{F}_j)$.
\end{restatable}

\begin{proof}
From Lemma~\ref{thm:correct}, the first adversarial facet deleted from $\pq$ is the nearest one to $x$, and if that happens, \padv terminates. It is implied by Theorem~\ref{thm:geocert2} that the facets are always deleted from $\pq$ in an ascending order of their distance to $x$. Combining these two facts, we can conclude that the distance of any facets deleted before the adversarial one is always smaller than $\epsilon^*$.
\end{proof}

\section{\padv Performance Optimization} \label{ap:opt}

We introduce a total of four performance optimizations to speed up the computation of \padv. 
We have explained the first two in Section~\ref{ssec:scale} and will describe all of them here in more details.

\subsection{Pruning Distant Facets}
This was described in the main text.
So here, we only provide examples of where the pruning can occur to remove unnecessary facet computation.
First, before computing the distance between $x$ and a facet, as proposed in  Section~\ref{ssec:distance_compute}, we can first use $\epsilon$ to filter unnecessary computation. Specifically, we can compute the orthogonal projection of $x$ onto the bisector implied by the facet. 
If the distance to the bisector, which is  a lower bound on the distance to the facet, is larger than $\epsilon$, then this facet can be safely discarded.

Even if we proceed with the distance computation, we can still use $\epsilon$ to terminate the optimization in Eqn.~\eqref{eq:distance} early. 
Specifically, if the dual objective of Eqn.~\eqref{eq:distance} surpasses $\epsilon^2$, we can terminate the solver and discard this facet since, from strong duality, the dual objective is a lower bound of the primal objective, i.e. the (squared) distance between $x$ and this facet.

\subsection{Rethinking the Initialization of $\epsilon$}
Recall that Line~1 of Algorithm~\ref{algo:padv} initializes $\epsilon$ to $\infty$. 
Given the upgraded role of $\epsilon$ in the previous paragraph, it is clear that a non-simplistic initialization would filter out more unnecessary computation early on and, thus, scale the overall performance.
A natural choice is to pick one of the baseline attacks for this purpose. 
The closer this adversarial distance is to the optimal one, the more computation we are likely to save by the first performance optimization, (I) Pruning Distant Facets. 

However, there is a trade-off between the tightness of the estimates and its computation time. 
Using an expensive attack to initialize $\epsilon$ can be a huge overhead that increases the total runtime rather than reduces it.
For our experiments, we run \sit to initialize $\epsilon$ since it yields a reasonable and is significantly faster than \yang and \wang.

\subsection{Exploiting the Sparsity of Solutions}
Solving a typical quadratic program has a complexity of $O(poly(n,d,k))$, but fortunately, this problem can be solved very efficiently in its dual form. \citet{wang19primal} show that solving Eqn.~\eqref{eq:distance} via greedy coordinate ascent (GCA) is much faster than using a standard off-the-shelf solver as it is able to exploit the sparsity in the solution. More details about this speedup can be found in Appendix~\ref{ap:dual}.

\subsection{Setting a Time Limit} 
To ensure that \padv terminates in a reasonable time even when no adversarial facet has been deleted from $\pq$, we set a time limit as a termination criterion. In this case, lower and upper bounds of $\epsilon^*$ are returned instead.
Note that the approximate version of \yang and \wang also terminates early by setting the maximum number of adversarial cells to search through instead of a limit on the runtime.

\section{Distance Computation with Greedy Coordinate Ascent} \label{ap:dual}

We first restate the distance computation from Eqn.~\eqref{eq:distance}:
\begin{align*}
    \min_{z}& \quad \norm{z - x}_2^2 \\
    \text{s.t.}& \quad Az \le b
\end{align*}

Note that without loss of generality, the equality constraint $\inner{z, \hat{a}} = \hat{b}$ can be subsumed by the inequality. 
Now we provide the dual form of Eqn.~\eqref{eq:distance}:
\begin{align*}
    \max_{\lambda}& \quad g(\lambda) \coloneqq -\frac{1}{2}\lambda^\top AA^\top \lambda + \lambda^\top (Ax - b) \\
    \text{s.t.}& \quad \lambda \ge 0
\end{align*}

In the case that the primal and the dual problems are feasible, we know that strong duality holds because the objective is convex quadratic, and the constraints are affine~\citep{boyd}. Thus, by setting the derivative of the Lagrangian to zero, we have that $z^* = x + A^\top \lambda^*$.

According to the complementary slackness from the KKT conditions, we know that $\lambda^*_i \ne 0$ if and only if $\inner{a_i, z^*} = b_i$, which geometrically corresponds to $z^*$ lying on the $i$-th bisector associated with $a_i$ and $b_i$. Intuitively, it is unlikely that $z^*$ lies on an intersection of many bisectors. Hence, there should be very few indices $i$ such that $\lambda^*_i \ne 0$. This is the condition that makes solving the dual problem with greedy coordinate ascent very fast~\citep{wang19primal}.

Greedy coordinate ascent (or descent) optimizes the variable only one coordinate per iteration, and there are multiple rules for choosing the coordinate at each iteration. Here, we follow \citet{wang19primal} and simply pick the $i$-th coordinate of $\lambda$ such that its projected gradient is the largest. We describe greedy coordinate ascent in Algorithm~\ref{algo:gca}. To avoid the full gradient computation at every iteration, we keep track and update it given that $\lambda$ only changes by one coordinate.

\begin{algorithm}[ht]
\small
\caption{Greedy Coordinate Ascent} \label{algo:gca}
	\KwData{Test point $(x,y)$, Voronoi cell described by $Az \le b$}
 	\KwResult{Projection of $x$ onto the Voronoi cell}
 	Initialize $\lambda \leftarrow \bm{0}$ \\
	\For{$t \in \{1,\dots,T\}$}{
	   $\nabla g(\lambda) \leftarrow -AA^\top \lambda + Ax - b$ \\
	   $j \leftarrow \argmax_i \abs{(\max\{\lambda + \nabla g(\lambda), 0\} - \lambda)_i}$ \\ 
	   $\lambda_j \leftarrow \max\left\{\lambda_j + \frac{\nabla g(\lambda)_j}{\norm{a_j}^2_2}, 0\right\}$ 
	}
	\Return{$z = x + A^\top \lambda$}\;
\end{algorithm}

\subsection{Details on Bisector Activeness Testing} 
We do a total of three checks to determine the feasibility: (i) check if the dual objective converges fast. 
When unbounded, the dual objective diverges or keeps increasing with a constant rate or faster. 
Additionally, we test whether the KKT conditions hold at the end of the optimization. 
Namely, (ii) check if the primal residual is zero, and (iii) check if the complementary slackness is satisfied. 
When all three checks pass, we conclude that the bisector is active. 
Otherwise, it is considered inactive, and there is no need to finish the distance computation or insert it to $\pq$.

\end{document}